\documentclass[lettersize,journal]{IEEEtran}
\usepackage{amsmath,amsfonts}
\usepackage{algorithmic}
\usepackage{algorithm}
\usepackage{xcolor}
\usepackage[table]{xcolor}
\usepackage{array}
\usepackage[caption=false,font=normalsize,labelfont=sf,textfont=sf]{subfig}
\usepackage{textcomp}
\usepackage{stfloats}
\usepackage{url}
\usepackage{verbatim}
\usepackage{graphicx}
\usepackage{cite}
\usepackage{multirow, multicol}

\usepackage{xr-hyper}
\usepackage{hyperref}
\begin{document}
\title{Neural Operator-enabled Topology-informed Evolutionary Strategy for PDE-Constrained Optimization}
\author{
Xiangming Huang$^{1}$,
Guannan Zhang$^{2}$,
Lu Lu$^{3}$,
and Raphaël Pestourie$^{1}$%
\thanks{$^{1}$Georgia Institute of Technology, School of Computational Science and Engineering, Atlanta, GA, USA.}%
\thanks{$^{2}$Oak Ridge National Laboratory, Oak Ridge, TN, USA.}%
\thanks{$^{3}$Yale University, New Haven, CT, USA.}%
}


\maketitle

\begin{center}
\begin{minipage}{0.95\linewidth}
\footnotesize

\noindent This is the accepted manuscript version for publication by IEEE.

\smallskip

\noindent \textcopyright~2026 IEEE. Personal use of this material is permitted.
Permission from IEEE must be obtained for all other uses, in any current or future media,
including reprinting/republishing this material for advertising or promotional purposes,
creating new collective works, for resale or redistribution to servers or lists, or reuse
of any copyrighted component of this work in other works.
\end{minipage}
\end{center}

\section{Abstract}

The inverse design of physical systems governed by partial differential equations is computationally demanding due to the high dimensionality and non-convexity of design spaces. Generative models for inverse design often lack robustness and transferability, whereas evolutionary strategies are robust but struggle in high-dimensional spaces. This paper introduces a Neural Operator–enabled Topology-informed Evolutionary Strategy (NOTES) that integrates dimensionality reduction, representation learning, and evolutionary optimization for efficient and transferable inverse design. NOTES couples a DeepONet-based neural operator with the Covariance Matrix Adaptation Evolution Strategy (CMA-ES) to perform global optimization in a compact latent space that encodes topology-aware priors while discovering high-performance designs for unseen operating conditions. Applied to nanophotonic beam-deflector inverse design governed by Maxwell’s equations, NOTES reduces the design dimensionality from 256 to 25 and consistently achieves over 95\% efficiency, outperforming CMA-ES, topology optimization, and other baselines. Applied to structural optimization, NOTES discovers designs that achieve compliance down to 246. By decoupling topology learning of a DeepONet from the governing physics in a PDE solver, NOTES provides a flexible and transferable framework for the inverse design of physical systems.

\section{Introduction}

We introduce a \emph{Neural Operator-enabled Topology-informed Evolutionary Strategy} (NOTES) that integrates implicit neural reparameterization and evolutionary optimization to provide an efficient, flexible, and transferable framework for inverse design of physical systems. The method unifies a DeepONet-based neural operator~\cite{lu2021learning} with the Covariance Matrix Adaptation Evolution Strategy (CMA-ES)~\cite{hansen2016cma} to perform global optimization in a compact latent space, which encodes topology-aware priors and reduces the number of optimization variables. Although NOTES does not embed the governing partial-differential equation (PDE) explicitly into the neural operator loss, the framework remains physics-informed through the construction of its training dataset. The DeepONet is trained exclusively on designs obtained from direct PDE-constrained optimization of the target physical systems, so the learned latent representation is biased toward geometries that encode physically meaningful high-performance features. In this sense, physical information enters NOTES indirectly through the data distribution rather than through a residual-based loss function\cite{RAISSI2019686}. In addition, DeepONet allows architectural inductive biases to be incorporated into the representation, illustrated here through binarization constraints and, in prior work, through hard enforcement of physical priors such as boundary conditions and invariance~\cite{lu2021physics}.

NOTES is particularly well suited for PDE-constrained inverse design problems where relevant high-performance designs are already available, allowing transferable geometric features to be learned from existing data. By combining representation learning with evolutionary optimization, the framework aims to make evolutionary strategies practical for high-dimensional topology optimization problems that are otherwise prohibitively expensive for direct search. In addition to improving optimization efficiency, NOTES provides a robust mechanism for consistently generating diverse high-performance designs across different physical settings.


\section{Prior Work}

\subsection{Evolutionary Strategy for Inverse Design}
CMA-ES~\cite{hansen2001completely, auger2005restart,Hansen2006,hansen2016cma,akimoto2016projection, loshchilov2018large} is a widely used derivative-free evolutionary strategy~\cite{slowik2020evolutionary} for nonconvex optimization because of its robustness and flexibility. However, its performance deteriorates rapidly in high-dimensional search spaces due to the curse of dimensionality~\cite{bellman2015applied}. Recent works~\cite{uchida2024covariance, kus2025gradient, guo2025advancing} therefore explore dimensionality reduction and latent-space optimization to improve scalability.


 \subsection{Representation Learning for Inverse Design}
 

Representation learning has been widely used to accelerate PDE-constrained inverse design~\cite{hinze2008optimization, woldseth2022use}. Several prior works~\cite{ma2019probabilistic,marzban2025hilab,liu2018generative} constructed various generative models for inverse design. Variational Autoencoders (VAEs) are widely used for this purpose across many applications~\cite{marzban2025hilab,singh2024deep,singh2025active,zhang2022inverse,ko2025inverse,ren2022invertible}. However, controlling inductive bias in VAEs remains challenging because it can conflict with the inherent bias introduced by the variational formulation~\cite{miao2021incorporating,yacoby2020failure}. Our method follows the line of work that couples neural representations with evolutionary optimization~\cite{guo2018indirect,kus2025gradient}.

\subsection{Neural Operator}

 Although DeepONet has been widely used in topology optimization and PDE learning~\cite{lu2021learning,jin2022mionet,lu2022comprehensive,jiao2025one,jiao2024solving,zhu2023reliable,yin2024scalable}, it is typically employed as a surrogate PDE solver that maps PDE coefficients to solution fields. In contrast, our framework uses DeepONet as a nonlinear topology representation that maps a low-dimensional latent space to the design topology. The latent space captures transferable geometric features from high-performance structures, enabling compact and reusable design representations. Our approach is further inspired by prior studies on binarization bias in neural networks~\cite{kou2025efficient}, implicit neural reparameterization~\cite{hoyer2019neural, joglekar2024generative}, transferability of neural operators~\cite{zhu2023reliable}, and periodicity enforcement in DeepONet~\cite{lu2022multifidelity}.


\begin{figure}[!t]
\centering
\includegraphics[width=3.25in]{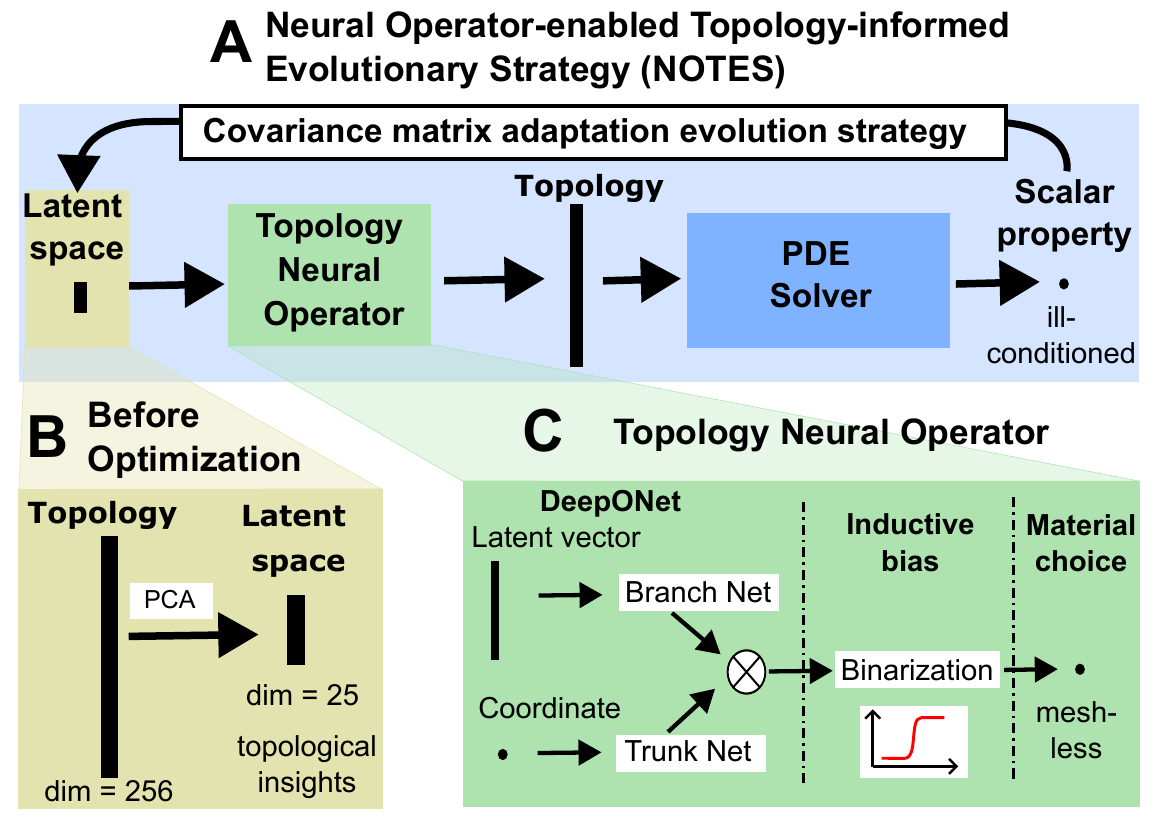}
\caption{(A) Overview of the neural operator-enabled topology-informed evolutionary algorithm. Using a latent space as the search space, topologies are generated by a neural operator. The target property of the generated topology is evaluated by a PDE solver. The CMA-ES algorithm is applied in the latent space to maximize the property. (B) Before the optimization, PCA reduces the high-dimensional design space to a compact latent space. The latent space captures key geometric features and provides a compact representation of the design space. (C) A neural operator, based on DeepONet~\cite{lu2021learning} is trained to map latent vectors to their corresponding design topologies. The architecture and loss function of the neural operator encourage the generated design to satisfy binarization constraints.}
\label{method}
\end{figure}

\begin{figure}[!t]
\centering
\includegraphics[width=3.25in]{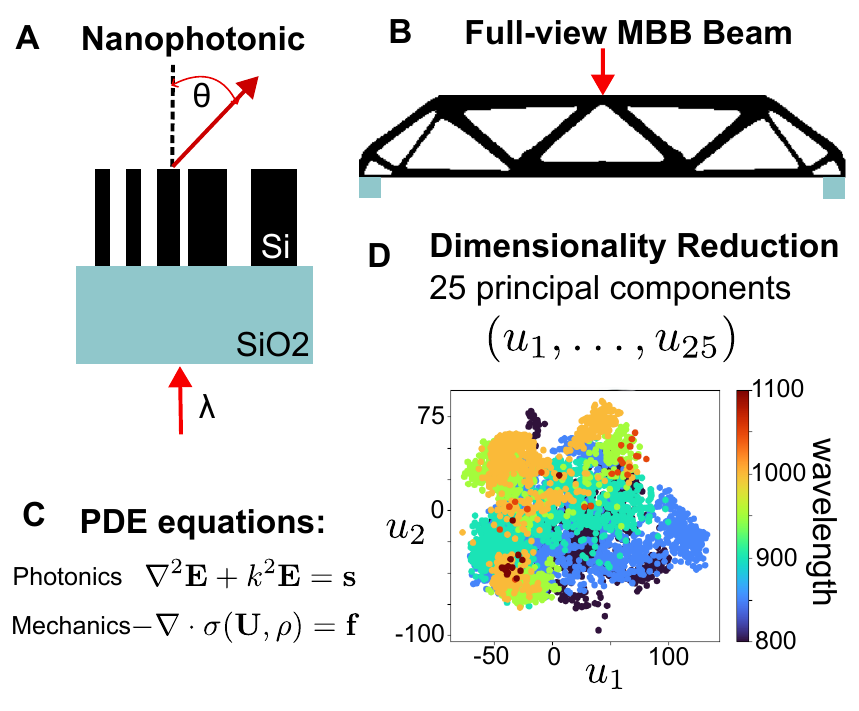}
\caption{Illustration of the applications and dimensionality reduction framework used in NOTES. (A) Side view of the 1D nanophotonic beam deflector. An incident TM-polarized plane wave with wavelength $\lambda$ propagates from the bottom, and the normalized transmitted power in the $\theta$ direction is computed as the deflection efficiency. (B) Full-view of the 2D Messerschmitt--Bölkow--Blohm (MBB) beam for structural optimization. A downward force is applied at the center, while both lower corners are supported. Using symmetry, only half of the domain is optimized. The objective is to minimize the elastic compliance (elastic potential energy) of the structure. (C) In both applications, the target property is a scalar quantity evaluated by solving the corresponding PDEs shown in the figure. (D) PCA is used for dimensionality reduction. We illustrate with the projection of 6129 high-performance photonics designs onto the first two principal components that reveals clustering behavior with respect to the incident wavelength, indicating that PCA captures meaningful geometric features of the design manifold.}
\label{Application}
\end{figure}

The remainder of this manuscript is organized as follows. Section~\ref{sec:method} presents the overall NOTES framework and optimization procedure. Section~\ref{sec:application} introduces two PDE-constrained inverse design applications used to evaluate the proposed method. Section~\ref{sec:results} presents comparisons with multiple baselines and ablation studies, including evaluations of CMA-ES as the optimizer, DeepONet as the nonlinear decoder, transferability across PDE settings, robustness to training randomness, and the importance of physics-guided training data.

\section{Method}
\label{sec:method}

\subsection{Schematic Overview}

NOTES addresses topology optimization with PDE constraints~\cite{hinze2008optimization}. Such optimization problems are often formulated as minimization problems. When a higher value of a figure of merit (FOM) is desired, the objective is reformulated as minimizing the negative of the FOM:

\begin{align}\label{eq:PDEopt}
    \min_{\mathbf{v} \in \mathcal{C}}
    -\mathrm{FOM}(\mathbf{x};\mathbf{v})
    \quad \text{s.t.} \quad
    \mathcal{F}(\mathbf{x};\mathbf{v}) = 0
\end{align}

where $\mathrm{FOM}$ is a scalar objective function, $\mathcal{F}$ denotes the PDE operator, and $\mathcal{C}$ represents the feasible space. Additional constraints may be added to enforce manufacturability~\cite{chen2024validation}. $\mathbf{x}$ is the PDE solution, and $\mathbf{v}$ is the design variable. Solving for the optimal design $\mathbf{v}^*$ is challenging because the objective is typically nonconvex and defined over a high-dimensional design space. In addition, evaluating the objective requires repeatedly solving the underlying PDE system, which is computationally expensive.

NOTES consists of two stages: training and optimization. During the training stage, we construct a latent representation from high-performance designs using Principal Component Analysis (PCA) and train a neural operator to map latent vectors to design topologies. During the optimization stage, illustrated in Figure~\ref{method}\textbf{A}, the trained neural operator acts as a geometric reparameterization of the design space, while CMA-ES performs optimization over the learned latent representation.

\subsection{Dimensionality Reduction}

We construct a low-dimensional latent space from the original design space using dimensionality reduction and train a neural network decoder that maps latent vectors to design topologies:
\begin{align}
\mathbf{v} = \mathcal{NN}(\mathbf{u})
,\quad \quad
\mathrm{dim}(\mathbf{u})=n\ll \mathrm{dim}(\mathbf{v})=N
\end{align}

As illustrated in Figure~\ref{method}\textbf{B}, the latent space captures key topological features while significantly reducing the dimensionality of the optimization problem. The PDE-constrained optimization problem is therefore reformulated as
\begin{equation}
\begin{aligned}
\min_{\mathbf{u\in\mathbb{R}^{n}}} \quad & -\mathrm{FOM}(\mathbf{x}; \mathcal{NN}(\mathbf{u})) \\
\text{s.t.} \quad 
& \mathcal{F}(\mathbf{x}; \mathcal{NN}(\mathbf{u})) = 0, \\
& \mathcal{NN}(\mathbf{u}) \in\mathbb{R}^{N}.
\end{aligned}
\end{equation}

Optimizing in the latent space mitigates the curse of dimensionality~\cite{bellman2015applied} and improves the scalability of evolutionary optimization~\cite{liu2024large}.

\subsection{Representation Learning}

We train a DeepONet-based neural operator that maps latent vectors and density-function coordinates to design topologies (Figure~\ref{method}\textbf{C}). Using this neural representation, we reformulate the optimization problem in the latent space:

\begin{equation}
\label{eq:latent_opt}
\begin{aligned}
\min_{\mathbf{u}\in\mathbb{R}^{n}}
\quad &
-\mathrm{FOM}(\mathbf{x}; \hat{\mathbf{v}})
\\
\text{s.t.}\quad &
\mathcal{F}(\mathbf{x}; \hat{\mathbf{v}})=0,
\\
&
\hat{\mathbf{v}} \in\mathbb{R}^{N},
\\
&
\hat{\mathbf{v}}
=
\sigma\!\left(
G(\mathbf{u}, \mathbf{y});
\hat{\eta}, \hat{\beta}
\right).
\end{aligned}
\end{equation}

Here, $G(\mathbf{u}, \mathbf{y})$ denotes the DeepONet used to output the design vector $\hat{\mathbf{v}}$. Its architecture consists of two sub-networks: the branch net $g(.)$, which encodes the latent vector $\mathbf{u}=(u_1, \dots, u_{n})$, and the trunk net $f(.)$, which encodes the density-function coordinates $\mathbf{y}$ in the design. To facilitate training, $\mathbf{y}$ is normalized between 0 and 1. Formally, DeepONet can be expressed as:

\begin{align}
    G(\mathbf{u}, \mathbf{y}) =  \underbrace{g(u_1, \dots, u_{n})}_{\mbox{branch}}\odot
    \underbrace{f(\mathbf{y})}_{\mbox{trunk}},
\end{align}
where $\odot$ denotes the component-by-component (Hadamard) product. Design topologies are parameterized as binary sequences because intermediate material states are nonphysical. To encourage binarized outputs, we apply a sigmoid transformation at the output layer of DeepONet~\cite{kou2025efficient}:

\begin{align}
    \sigma(\hat u;\beta,\eta) &= \frac{1}{1+\exp(-\beta \hat u-\eta)} \label{custom_sig}
\end{align}

where $\beta$ controls the degree of binarization from a threshold $\eta$. The model is trained using binary cross-entropy loss together with a binarization regularization term:

\begin{align}
\mathcal{L}_{BCE}&=-\frac{1}{N}\sum_{i=1}^N\left[\mathbf{v}^T \log \hat{\mathbf{v}}+(1-\mathbf{v})^T \log(1-\hat{\mathbf{v}})\right]\\
\mathcal{L}_{bias} &= (\mathbf{1}_{N} - \hat{\mathbf{v}})^T\hat{\mathbf{v}} \label{bias}\\
\mathcal{L} &= \mathcal{L}_{BCE}+ \alpha\mathcal{L}_{bias}\label{loss}
\end{align}

Here, $\alpha$ controls the contribution of the binarization regularization term and $\mathbf{1}_{N}$ denotes a N-dimensional vector of ones. Although binarization is enforced as a soft constraint, progressively increasing $\beta$ during training yields increasingly binarized topologies. Note that fabrication-aware priors are incorporated implicitly through the neural representation and the distribution of high-performance training designs rather than through the final optimization objective.

\subsection{Optimization Over Latent Space}

Given a trained neural operator, CMA-ES is applied to solve the optimization problem defined in Eq.~\ref{eq:latent_opt}. The latent vector $\mathbf{u}$ is initialized through random sampling in the latent space, with the sampling scale determined from the latent-coordinate distribution of the training data. During optimization, each generated topology is evaluated using a PDE solver to compute the scalar objective function. The PDE solver ensures accurate physics evaluation and is the dominant computational cost in NOTES.

\section{Application}\label{sec:application}


\subsection{Problem Setup}

To demonstrate both the effectiveness and applicability of NOTES across physical domains, we evaluate the proposed framework on two PDE-constrained inverse design applications with fundamentally different governing physics and geometric structures. The first is a one-dimensional nanophotonic inverse design, where the objective is to generate high-efficiency metagrating patterns for prescribed incident wavelengths and deflection angles. The second is a two-dimensional structural inverse design, where the objective is to generate low-compliance load-bearing structures under prescribed design resolutions and material volume constraints. The nanophotonic application serves as the primary benchmark throughout the main manuscript because it provides a well-established inverse design setting with strong competing baselines, while the structural mechanics application is used to further verify that NOTES is transferable across different PDEs or physical domains and effective in a more complex design space.


We perform the inverse design of a beam deflector, following settings similar to Ref.~\cite{jiang2020simulator}. The objective is to maximize the deflection efficiency (the ratio of the electromagnetic power deflected to the incident light power) of the $+1$ diffraction order in silicon nanoridge metagratings. The grating thickness is fixed at 325 nm, and the incident light is TM-polarized. The grating period equals the incident wavelength divided by the sine of the deflection angle in magnitude and is discretized into 256 equal ridges, each assigned one of two material states.
Figure~\ref{Application} \textbf{A} illustrates a single grating period. Each period is parameterized as a 256-dimensional binary vector, where 1 denotes silicon and 0 denotes air. A differentiable PDE solver, Meent\cite{kim2024meent}, is then used to solve the PDE in Figure~\ref{Application} \textbf{C} and to evaluate the objective, with the incident wavelength $\lambda$ and deflection angle $\theta$ as parameters:

\begin{align}\label{eq:setup}
    \min_{\mathbf{v}\in\{0,1\}^{256}} -P(\mathbf{v}; \lambda,\theta)
\end{align}
The deflection efficiency is a percentage, denoted as $P$ and dependent on $E$ (Fig.~\ref{Application}\textbf{C}). The permittivity profile, denoted as $\mathbf{v}$, is a binary vector. The objective is to find the optimal $\mathbf{v}^*$ that maximizes $P$. As in Eq.~\ref{eq:latent_opt}, we use the optimization formulation with our DeepONet representation on a latent space of dimension $n=25$.

The second application considers a two-dimensional structural topology optimization problem. We apply NOTES to the classical Messerschmitt--Bölkow--Blohm (MBB) beam benchmark on a design domain of size $192 \times 64$, resulting in 12,288 pixel-level design variables, illustrated in Fig.~\ref{Application}\textbf{B}. The overall problem setup and finite-element implementation follow the framework in Ref.~\cite{hoyer2019neural}. 
The discrete compliance minimization problem is given by

\begin{equation}
\begin{aligned}
\min_{\mathbf{v}\in\mathbb{R}^{192\times64}} \quad & U^{T} K U \\
\text{subject to} \quad 
& K U = F, \\
& V(\mathbf{v}) = V_{0}, \\
& 0 \le \mathbf{v}_{ij} \le 1, \quad \forall\, (i,j),
\end{aligned}
\label{eq:2D-objective}
\end{equation}
where $\mathbf{v}_{ij}$ denotes the physical density at each pixel, $K(\mathbf{v})$ is the global stiffness matrix, $F$ is the external load vector, and $U(K,F)$ is the corresponding displacement vector obtained from the finite-element equilibrium equation. The total material volume $V(\mathbf{v})$ is constrained to a predefined volume fraction $V_0$.

The following constrained sigmoid transformation is applied after the forward pass in the DeepONet to enforce the volume and density constraints in Equation~\ref{eq:2D-objective}:
\begin{equation}
\begin{aligned}
\tilde{\mathbf{v}}_{ij} &= \frac{1}{1+\exp[-\hat{\mathbf{v}}_{ij}-b(\hat{\mathbf{v}}, V_0)]}, \\
\text{such that} \quad & V(\tilde{\mathbf{v}})=V_0,
\end{aligned}
\label{eq:vol-frac-enf}
\end{equation}
where $b(\hat{\mathbf{v}},V_0)$ is a scalar bias term determined by binary search so that the transformed output field satisfies the predefined volume fraction. Finally, the physical density field $\mathbf{v}$ is obtained by applying a cone filter with a radius of 2 to $\tilde{\mathbf{v}}$, which smooths isolated small features as in Ref.~\cite{hoyer2019neural}. As in Eq.~\ref{eq:latent_opt}, we use the optimization formulation with our DeepONet representation on a latent space of dimension $n=60$.

\subsection{Training Data and Latent Space}


For the nanophotonic beam deflector application, we use the publicly available MetaNet benchmark dataset~\cite{jiang2020metanet}, which contains high-performance grating designs generated by GLOnet and adjoint-based topology optimization~\cite{jiang2020simulator} under multiple (wavelength, deflection angle) operating conditions. From this dataset, we retain only designs whose deflection efficiency exceeds $0.9$ for their corresponding operating pair $(\lambda,\theta)$.

A practical difficulty is that the deflection efficiency is not uniformly distributed across all operating conditions. Several wavelength--angle pairs contain no high-efficiency designs. For instance, under $(\lambda,\theta)=(1100\text{ nm},40^\circ)$, no design with efficiency above $0.9$ is available in the database, and the best designs selected from other operating conditions achieve only about $0.75$ efficiency. Therefore, the neural operator cannot rely solely on direct memorization of condition-specific optimal patterns. Instead, it must learn transferable topological features from high-efficiency designs to generate new candidate structures for previously underrepresented operating conditions.


To construct a latent space for DeepONet in photonics, we applied PCA~\cite{bishop2006pattern} to the dataset of 6129 high-performance designs spanning different wavelengths and deflection angles, following the procedure outlined in Metanet~\cite{jiang2020metanet} that conserves the translational invariance of the periodic grating. The physical parameters ($\lambda, \theta$) are not directly encoded in the latent space, only indirectly influencing the latent representation through their corresponding high-performance geometries in the dataset. The distribution of data points along the first two principal components is visualized in Figure~\ref{Application}\textbf{D}, which reveals distinct clustering by wavelength. The visualization indicates that the latent space captures distinct geometric features associated with different wavelengths and deflection angles. The first 25 principal components explain more than 90\% of the total variation and are chosen as our low-dimensional latent space. With this latent space, DeepONet learns a 25-dimensional representation that generates 256-dimensional designs.

For the MBB beam application, the dataset is generated using density-based topology optimization with the L-BFGS optimizer. The designs are generated with a prescribed volume fraction of $0.4$. To improve diversity in the generated dataset, each new optimization is initialized from the previous design after applying erosion operations with varying rates. The resulting dataset has a median compliance of approximately 270 (SI section 2.1 \& 2.2). We retain 857 optimized designs as base samples and further augment the dataset using additional erosion and dilation operations (Figure S8 in SI). For each base design, two augmented variants are generated. After augmentation, the complete dataset contains 2571 designs, which are randomly split into 2056 training samples and 515 testing samples. To obtain the latent space for the MBB beam application, we fit PCA on the training samples and retain the first 60 principal components, which explain approximately $83\%$ of the variance in the training dataset (Figure S9 in SI).

\subsection{Training and Optimization}

The DeepONet implementation is provided by DeepXDE~\cite{lu2021deepxde}. Using a $60/40$ train/test split, the model is trained on 3677 samples for the photonic case. Both the branch and trunk networks contain three hidden layers with 60 neurons per layer, resulting in approximately 20k trainable parameters. A sigmoid activation is applied at the output with an adaptive scale factor and a threshold initialized to the average value of the training data. The network is trained using the ADAM optimizer with learning rate $10^{-3}$ for 400,000 epochs. The final validation MSE is approximately 0.06 (Figure S2 in SI). The trained DeepONet achieves high reconstruction quality (Figure S3 in SI).

The PDE solver is provided by Meent~\cite{kim2024meent}, which implements differentiable rigorous coupled-wave analysis (RCWA). This enables both gradient-based optimization and evolutionary optimization in the learned latent space. CMA-ES is implemented using pycma~\cite{hansen2024cma}. 

Preliminary optimization results suggest that including a binarization term in the loss function is helpful. Approximately $20\%$ of the generated designs remain partially non-binary without binarization bias, while incorporating a binarization bias reduces this fraction to about $3\%$.

The latent vector is initialized randomly within the PCA coordinate range observed in the training dataset. CMA-ES is applied with a population size of 20 and a maximum of 100 iterations. Different initial sampling radii are used for the baseline CMA-ES and for NOTES ($\sigma_0=10$ and $\sigma_0=1000$, respectively), resulting from hyperparameter tuning. The initial standard deviation for NOTES is relatively large because the PCA latent coordinates are not normalized.


For the MBB beam application, the number of parameters in the DeepONet is increased to 1.6 million because of the output complexity (SI section 2.3). The training procedure follows the same setup as the 1D photonic case, including the use of a custom sigmoid activation and a binarization bias in the loss function. However, unlike the photonic application, the optimizer enforces the volume constraint strictly instead of the DeepONet. The trained DeepONet achieves high reconstruction quality for the 2D mechanical case (Figure S10 in SI).

\begin{figure}[!t]
\centering
\includegraphics[width=3.25in]{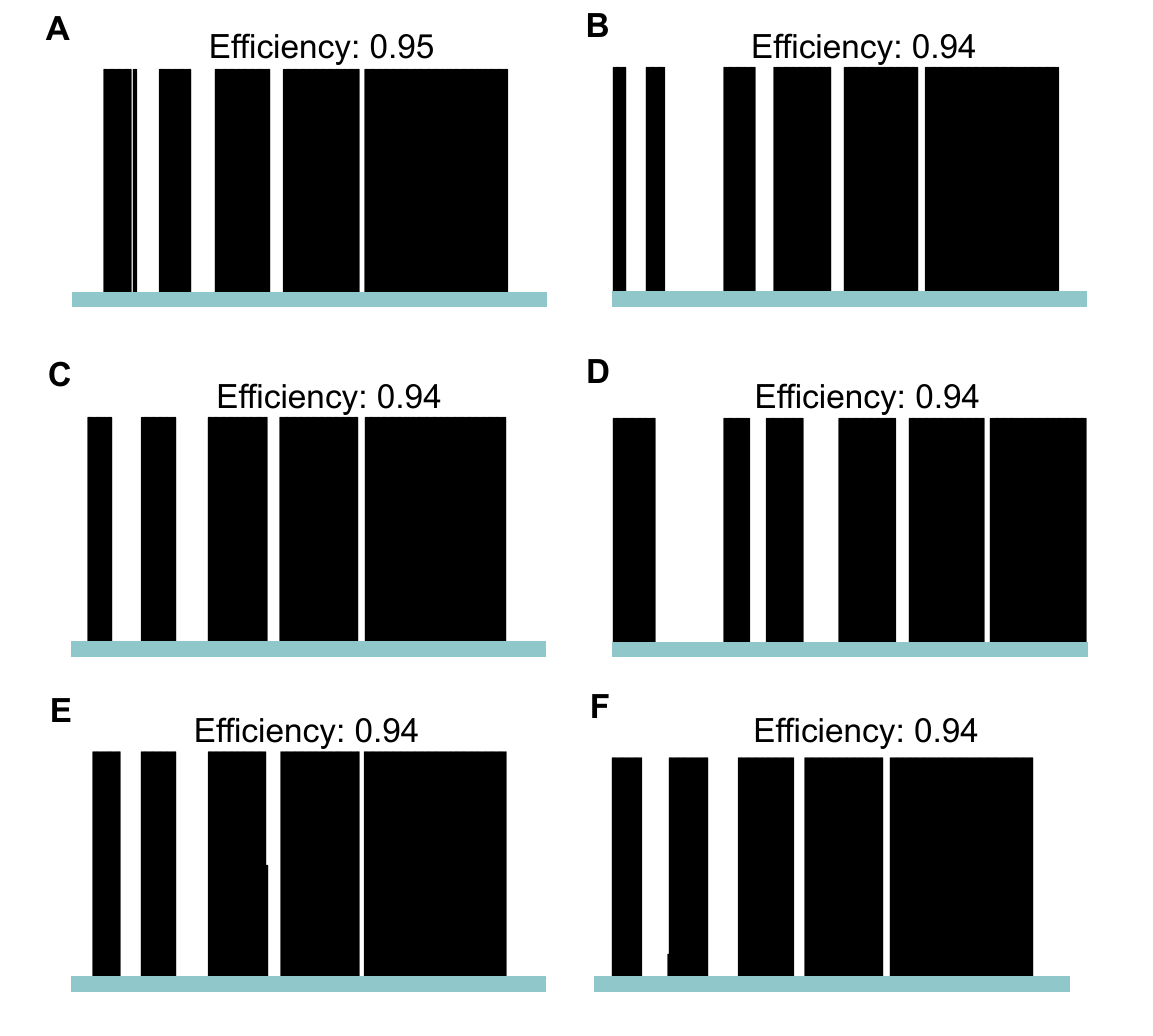}
\caption{(A-F) qualitatively present six optimized metagrating patterns generated by NOTES for the 1100 nm wavelength and 60° deflection angle. Although the best design in the training data achieves an efficiency equal to 0.9 for this setting, the method consistently produces results with higher efficiency. This shows the ability of NOTES to discover designs beyond the training data. Most generated designs satisfy binarization constraints. (E, F) show designs with minor imperfections. These cases are rare ($3\%)$ across all operating conditions. The aspect ratios of the six designs are 65.52, 16.38, 16.38, 16.38, 16.38, and 21.84, respectively. High-aspect-ratio structures are uncommon, with 95\% of the generated designs having aspect ratios below 33.}
\label{Results}
\end{figure}

\begin{figure}[!t]
\centering
\includegraphics[width=3.25 in]{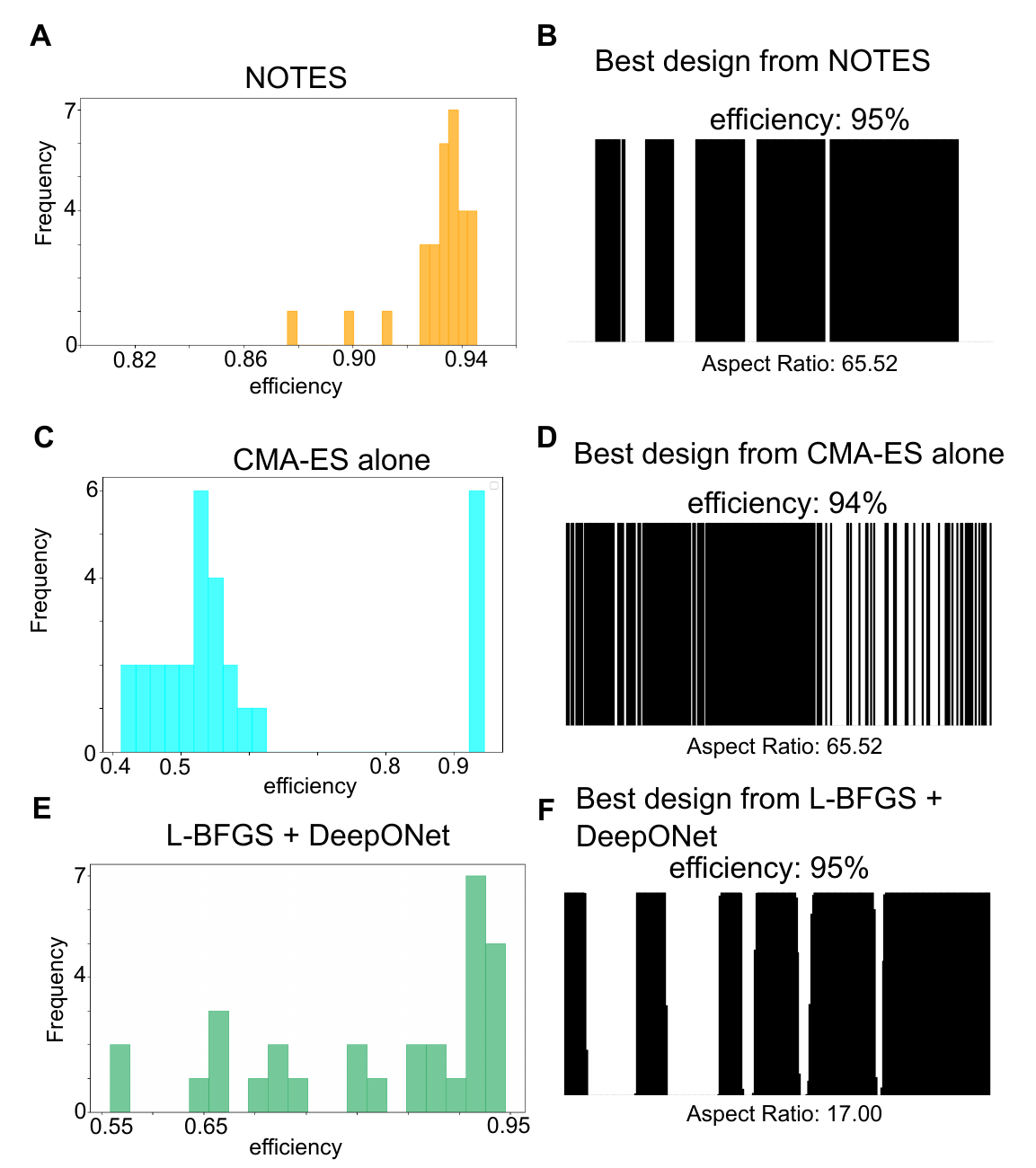}
\caption{
This figure compares NOTES with the standard CMA-ES and L-BFGS + DeepONet for the 1100~nm wavelength and 60° deflection-angle setting. Comparing panels (A), (C), and (E), NOTES exhibits substantially lower variance than the other two methods, with most generated designs achieving higher than 92\% efficiency. The aspect ratios of designs (B), (D), and (F) are 65.52, 65.52, and 17.00, respectively, with panel (F) containing intermediate nonphysical material values. 
}
\label{Comparison}
\end{figure}

\begin{figure}[!t]
\centering
\includegraphics[width=3.25in]{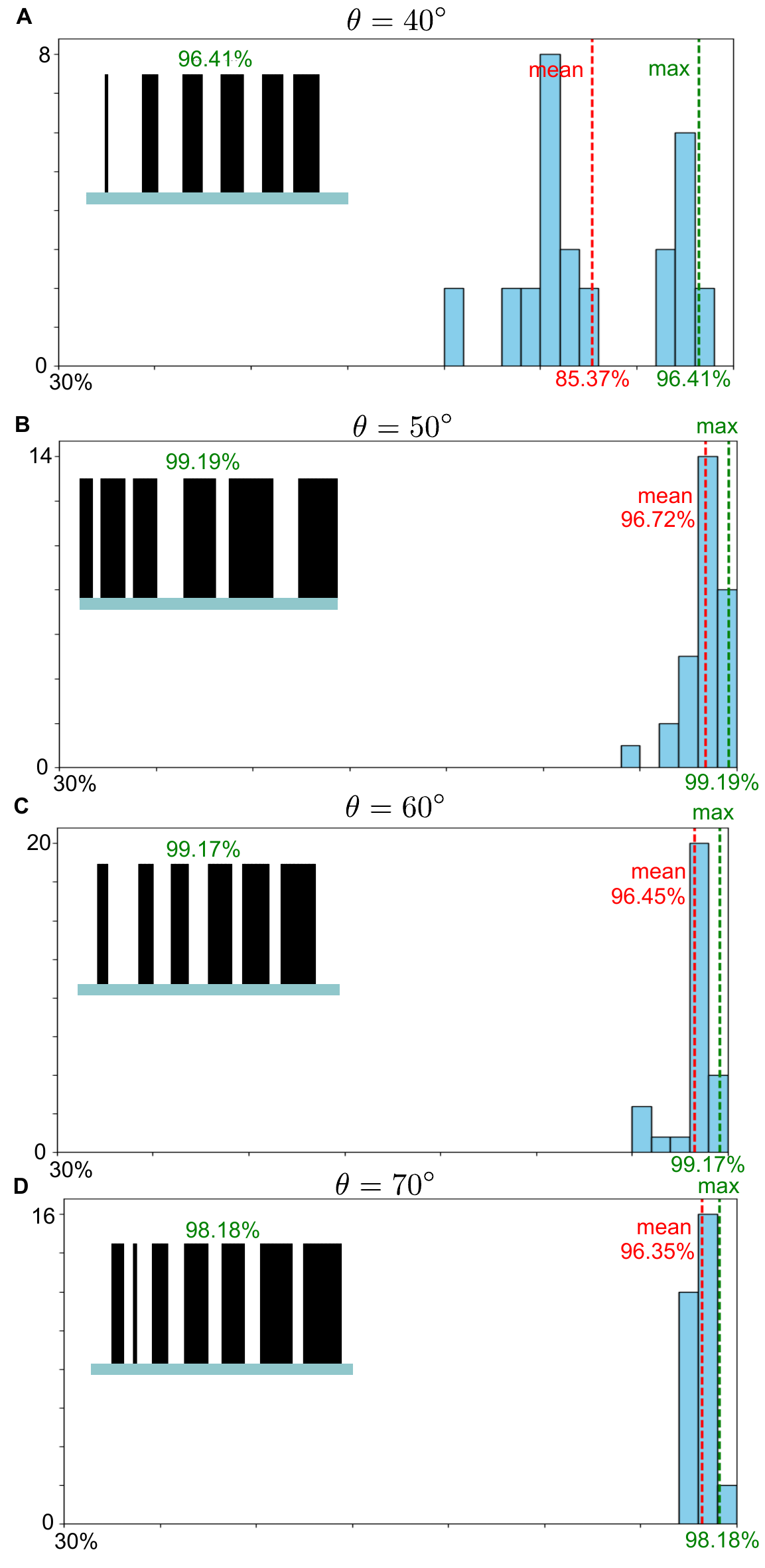}
\caption{Out-of-distribution performance of NOTES for visible-wavelength metagrating optimization. Panels (A–D) show optimized metagrating patterns at a wavelength of 530nm, which lies far outside the training range whose lower bound is 900nm. The mean (red dashed line) and maximum (green dashed line) values of the distribution are indicated in the histogram. The target deflection angles are (A) 40°, (B) 50°, (C) 60°, and (D) 70°.}
\label{Transfer learning}
\end{figure}

\begin{figure}[!t]
\centering
\includegraphics[width=3.25 in]{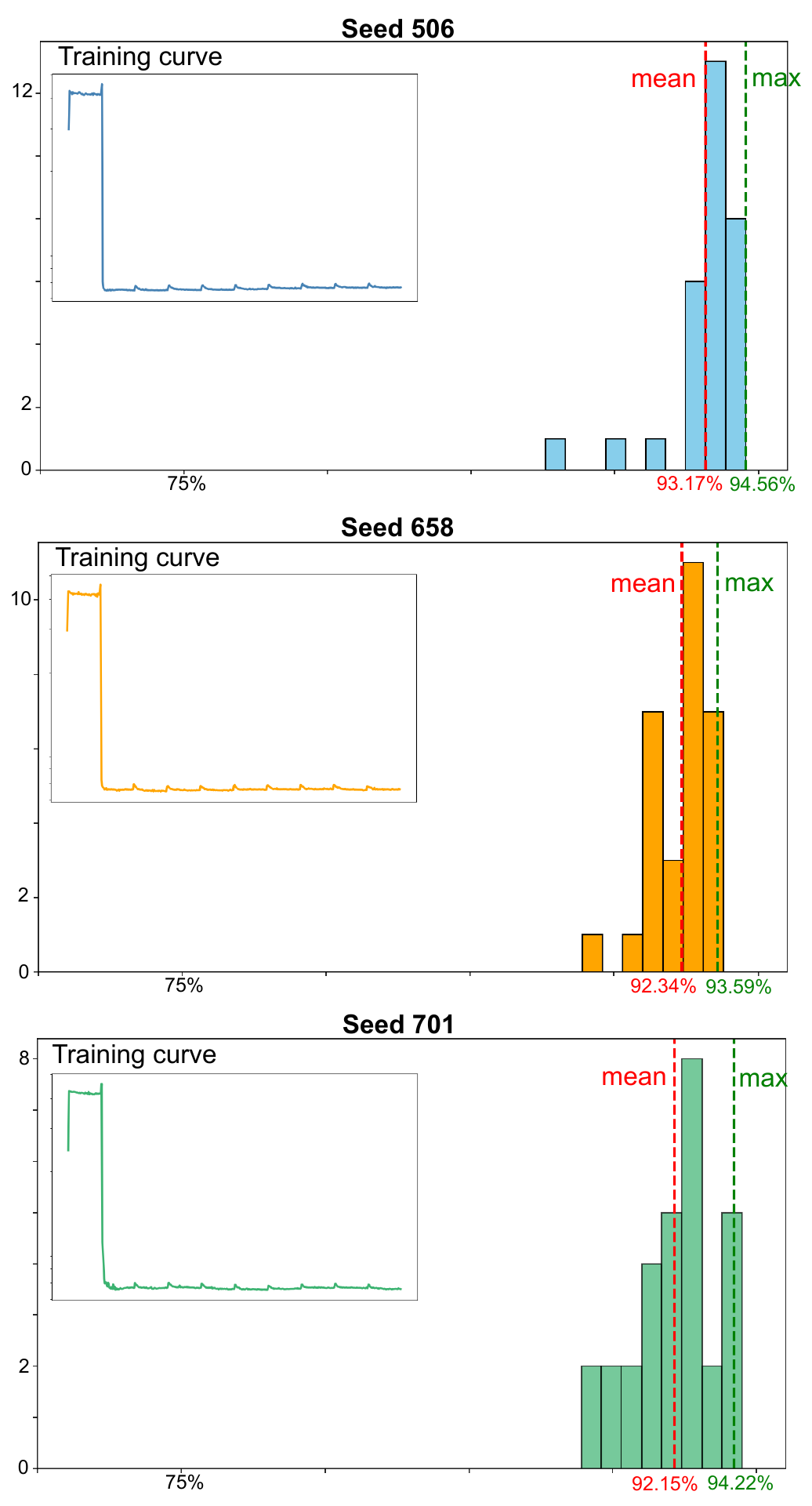}
\caption{
This figure demonstrates the robustness of NOTES when the neural operator is trained with different random seeds. The mean (red dashed line) and maximum (green dashed line) values of the distribution are indicated in the histogram. The corresponding neural operator training curves are shown in insets.}
\label{Stability}
\end{figure}

\section{Results}\label{sec:results}






\subsection{Baselines Configuration}

For the nanophotonic application, we compare NOTES against four baselines: topology optimization, GLOnet~\cite{jiang2020simulator}, CMA-ES, and L-BFGS + DeepONet. Topology optimization performs direct gradient-based optimization of the high-dimensional design. GLOnet~\cite{jiang2020simulator} is a sampling-based inverse-design framework that previously achieved state-of-the-art performance for this application. Standard CMA-ES performs direct evolutionary optimization in the 256-dimensional topology. Finally, L-BFGS + DeepONet applies the L-BFGS optimizer in the same latent space and neural representation as NOTES, differing only in the choice of optimizer. For the mechanical application, the baselines are L-BFGS-based topology optimization in high-dimensional space and the L-BFGS + DeepONet baseline. Due to the 12,288-dimensional feasible space, direct CMA-ES could not achieve comparable performance. In both applications, we also compare NOTES to optimizing on principal components directly, without using the DeepONet representation.

\subsection{Performance Summary} 
\label{subsec:compare-baselines}

For an incident wavelength of $\lambda = 1100~\mathrm{nm}$ and deflection angle $\theta = 60^\circ$, Figure~\ref{Results} qualitatively presents six optimized metagrating patterns generated by NOTES. Panels (E) and (F) show rare cases with minor non-binarized regions, although most generated designs remain fully binary across all operating conditions. In nanophotonics, the aspect ratio, defined as the ratio between the device height and the minimum feature width, is an important indicator of fabrication difficulty, with larger values generally being harder to manufacture. The obtained designs exhibit aspect ratios of 65.52, 16.38, 16.38, 16.38, 16.38, and 21.84. Overall, $95\%$ of the designs generated by NOTES have aspect ratios below 33.

Figure~\ref{Comparison} presents the distribution of optimized designs for a wavelength of 1100~nm and a deflection angle of 60° for NOTES, CMA-ES, and L-BFGS + DeepONet. CMA-ES exhibits substantially higher variance in efficiency than NOTES. As shown in panels (A) and (C), most designs produced by NOTES converge to efficiencies higher than 92\%, whereas CMA-ES yields a much broader distribution, with most designs below 65\% and only a few outliers near 94\%. This behavior reflects the difficulty of applying evolutionary strategies directly in a high-dimensional design space. L-BFGS + DeepONet also exhibits higher variance than NOTES, with a left-skewed efficiency distribution, suggesting frequent convergence to poor local optima, analyzed in more detail in the Section ~\ref{subsec:optimizer}. These results indicate that CMA-ES is more effective than L-BFGS at identifying high-quality optima in the learned latent space.

From the results in Table~\ref{table:results}, among all methods, NOTES demonstrates the best overall performance in most cases. Even when it does not achieve the top result, its performance remains within 2\% of the best. NOTES consistently achieves the highest mean efficiencies and the lowest variance. Designs generated by NOTES also have a high probability of satisfying the binarization constraint. $95\%$ of the designs from NOTES have aspect ratios less than 33, and $97\%$ of the results are fully binary. By learning from existing data, it generates new designs with efficiencies that are, on average, 3\% higher than GLOnet~\cite{jiang2020simulator}. Although this improvement may appear modest, it is quite significant given that GLOnet~\cite{jiang2020simulator} has state-of-the-art performance with many designs close to the global optimum, 100\%. L-BFGS + DeepONet sometimes yields the highest-performing designs, but they are not necessarily binarized and, therefore, nonphysical. Baseline CMA-ES, which uses four times more iterations than NOTES, exhibits highly unstable performance, showing variance that ranges from the highest to nearly zero and efficiencies that vary from the best to the worst among all methods.

For the MBB beam application, we find that NOTES achieves a compliance of 246, demonstrating a $1.6\%$ improvement over L-BFGS-based topology optimization (SI section 2.4 \& Table S1). NOTES also reaches the same best performance as L-BFGS + DeepONet.

Regarding computational costs, solving the PDE is the most expensive part, largely dominating the DeepONet training costs. Each iteration, the evaluation of the objective function requires solving the PDE. We have found that NOTES has a faster convergence rate of up to one order of magnitude compared to direct CMA-ES (see Figure~S1 in SI), achieving a better tradeoff between performance and the number of PDE evaluations.

\subsection{L-BFGS + DeepONet vs NOTES}
\label{subsec:optimizer}

Table ~\ref{table:results} shows that L-BFGS + DeepONet exhibits substantially higher variance than NOTES in the photonic application. To understand the reason for such behavior, we compared the $L_2$ gradient norms with respect to the latent space parameters (Figure S4 in SI). With a 25-dimensional latent space, the average gradient norm for NOTES is $2 \times 10^{-4}$, whereas the average gradient norm for L-BFGS + DeepONet is $7.5 \times 10^{-3}$, both achieving designs that are near local optima. To further validate this observation, we applied additional gradient descent steps to the solutions obtained from L-BFGS + DeepONet. Although the average gradient norm was further reduced to $4 \times 10^{-3}$, the efficiency distribution remained nearly unchanged (Figure S5 in SI). However, the significantly higher variance observed for L-BFGS + DeepONet suggests that the optimization process frequently converges to poor local optima. We infer from these results that the larger variance of L-BFGS + DeepONet is primarily caused by convergence to poor local optima.

In contrast, for the MBB beam application, L-BFGS + DeepONet and NOTES achieve comparable performance and converge to similar designs (Figure S11 \& S16 in SI). These results indicate that the latent optimization landscape presents fewer poor local optima.

\subsection{DeepONet vs PCA decoder}
\label{subsec:decoder}

To examine whether DeepONet is necessary to achieve strong performance, we conduct additional comparative studies between a linear PCA-decoder coupled with CMA-ES and the proposed nonlinear DeepONet decoder used in NOTES across both applications. For fairness, the PCA reconstruction is employed with the same post-processing transformations as NOTES in each application. For the 1D nanophotonic beam deflector, PCA-decoder + CMA-ES performs comparably to NOTES and in some cases achieves slightly higher best efficiencies, with an average improvement of approximately 1\% under the same computational budget (Figure S6 in SI).

However, in the more geometrically complex 2D MBB beam, the PCA-decoder substantially underperforms NOTES (Figure S14 in SI). Under the same budget, NOTES achieves a best compliance of 246, whereas PCA-decoder + CMA-ES only achieves approximately 394, and improves to approximately 300 with twice the computational budget. This contrast reveals that the effectiveness of linear PCA reconstruction strongly depends on the degree of dimensional compression and the complexity of the feasible topology manifold. In the 1D case, reducing from 256 dimensions to a 25-dimensional latent space still preserves enough geometric information for linear reconstruction, whereas in the 2D case the compression from 12,288 dimensions to only 60 latent variables requires recovering a highly nonlinear mapping that PCA cannot adequately represent.

These results confirm that the performance gain of NOTES does not arise only from the PCA latent space, but also from the nonlinear expressive power of the DeepONet decoder, which reconstructs a more complex manifold of feasible high-performing PDE-constrained topologies.

\subsection{Reusability}
\label{subsec:transfer_learning}

The ability of NOTES to generate high-performance designs beyond the training distribution is demonstrated in Table~\ref{table:results}. During data selection, only designs with pre-computed/labeled efficiency $\geq0.9$ were retained from the MetaNet dataset~\cite{jiang2020metanet}. For the angle–wavelength combinations of $(1100\text{ nm},40^\circ)$, $(900\text{ nm},50^\circ)$, and $(1100\text{ nm},60^\circ)$, no design with pre-computed/labeled efficiency satisfying this threshold is available. However, after reevaluating designs in the training data, we find that for $(1100\text{ nm},40^\circ)$, the best design achieves an efficiency of 75\%, whereas NOTES produces a new design with 87\% efficiency. For $(900\text{ nm},50^\circ)$, the best design available in the training data achieves 98\% efficiency; NOTES discovers designs reaching 99\%, with smaller variance and a substantially faster convergence rate. For $(1100\text{ nm},40^\circ)$, the best design available in the training data achieves 90\% efficiency, while NOTES achieves 95\%, representing a 5\% gain. Beyond these three illustrations, NOTES consistently generates new designs that outperform the training data.

Figure~\ref{Transfer learning} further demonstrates the transferability of NOTES. By learning a compact neural reparameterization, NOTES accelerates optimization and generalizes effectively under unseen operating conditions. Panels (A–D) present metagrating optimizations at \(530\,\mathrm{nm}\) (green), far outside the training wavelength range, targeting deflection angles of \(40^\circ\), \(50^\circ\), \(60^\circ\), and \(70^\circ\). Despite operating at an out-of-distribution wavelength, the method yields patterns with near-perfect deflection efficiency (\(\approx 100\%\)). It is important to note that these are simulation results, in which the efficiencies are overestimated because visible-light absorption in silicon is neglected. In panel (A), the run-to-run variance is high, while our usual empirical findings align more closely with panels (B–D). This highlights the occasional instability but strong reusability of the trained operator.  From (A) to (D), the aspect ratios of those designs are low at 10.09, 15.03, 13.59, and 16.39 respectively, all of which remain within a reasonable fabrication range.

For the MBB beam application, we also find that the trained DeepONet can be used for different resolutions and volume constraints. For example, we apply the trained DeepONet to optimize designs at the 96×32 resolution with a volume constraint of 0.5 and at the 384×128 resolution with a volume constraint of 0.3. In both cases, the pretrained NOTES perform well compared to the results of pixel-LBFGS from previous work~\cite{hoyer2019neural}. At the 96×32 resolution, pretrained NOTES achieves a design that is within 2.4\% of the best L-BFGS design (Figure S13 in SI). At the 384×128 resolution, pretrained NOTES is within $1.8\%$ of the best L-BFGS design (Figure S12 in SI). Thanks to its meshless property, DeepONet seamlessly allows for different resolutions. These results demonstrate the transferability of the DeepONet representation.

\subsection{Robustness of NOTES}
\label{subsec:robustness}

The stability analysis demonstrates the robustness of NOTES when the neural operator is trained with different random seeds (Figure~\ref{Stability}). Despite variations in initializations, all neural operators produce nearly identical training curves. The resulting NOTES produce design distributions with similar shapes, means, and maximum efficiencies, indicating that the NOTES framework and induced latent space are robust to variations in neural-operator initialization.

\subsection{Comparison Between Randomly-sampled and Physics-informed Data}
\label{subsec:physics-informed comparison}

To further validate that the performance of NOTES arises from its physics-informed training data rather than from the DeepONet architecture alone, we conduct an ablation study on both the nanophotonic and MBB beam applications using DeepONets trained on designs sampled uniformly at random instead of topology-optimized designs. For the nanophotonic application, although both methods achieve similar maximum efficiencies, the randomly sampled approach exhibits significantly larger variance in performance and produces more designs with high aspect ratios (Figure S7 in SI). For the MBB beam application, NOTES trained on randomly-sampled data generates designs with compliance 735 (Figure S15 in SI), far worse than the compliance of 246 achieved by its physics-informed counterpart. These results demonstrate that the key advantage of NOTES does not arise only from latent-space dimensionality reduction or the DeepONet decoder, but also from the physically meaningful manifold induced by high-performance PDE-optimized training data, which embeds transferable topological priors essential for efficient downstream search.

\begin{table*}[htbp]
\centering
\renewcommand{\arraystretch}{1.2}
\begin{tabular}{c c c c c c c}
\hline
Angle / Wavelength & Metric & Topology Opt. & GLOnet & L-BFGS + DeepONet & CMA-ES & NOTES\\
(deg / nm) & & & & & & \\
\hline
40 / 900  & Mean & 68\% & 88\% & 69\% & 86\%& \cellcolor{green!25}93\%\\
          & Max  & 94\% & \cellcolor{blue!25}95\% & \cellcolor{blue!25}95\% & 94\%& \cellcolor{blue!25}95\%\\
          & Std  & 16\% & 8\% & 20\% & 16\%& \cellcolor{orange!25}2\%\\
\hline
40 / 1000 & Mean & 65\% & 69\% & 72\% & 66\%& \cellcolor{green!25}88\%\\
          & Max  & 90\% & \cellcolor{blue!25}92\%& 91\%& 86\%& 91\%\\
          & Std  & 14\% & 20\% & 16\% & 20\%& \cellcolor{orange!25}2\%\\
\hline
40 / 1100 & Mean & 61\% & 60\% & 74\% & \cellcolor{green!25}82\%& \cellcolor{green!25}82\%\\
          & Max  & 87\% & 86\% & 85\% & \cellcolor{blue!25}88\%& 87\%\\
          & Std  & 10\% & 15\% & 9\% & 6\%& \cellcolor{orange!25}2\%\\
\hline
50 / 900  & Mean & 64\% & 90\% & 81\% & 92\%& \cellcolor{green!25}98\%\\
          & Max  & 93\% & 98\% & \cellcolor{blue!25}99\% & 96\%& \cellcolor{blue!25}99\% \\
          & Std  & 16\% & 10\% & 17\% & 2\%& \cellcolor{orange!25}1\%\\
\hline
50 / 1000 & Mean & 55\% & 85\% & 81\% & 94\%& \cellcolor{green!25}96\%\\
          & Max  & 95\% & 96\% & \cellcolor{blue!25}97\% & 96\%& \cellcolor{blue!25}97\%\\
          & Std  & 16\% & 12\% & 17\% & 2\%& \cellcolor{orange!25}1\%\\
\hline
50 / 1100 & Mean & 49\% & 77\% & 82\% & 91\%& \cellcolor{green!25}93\%\\
          & Max  & 91\% & 91\% & \cellcolor{blue!25}98\% & 94\%& 96\%\\
          & Std  & 10\% & 11\% & 13\% & \cellcolor{orange!25}1\%& 2\%\\
\hline
60 / 900  & Mean & 59\% & 73\% & 86\% & \cellcolor{green!25}99\%& 98\%\\
          & Max  & 93\% & 97\% & \cellcolor{blue!25}99\% & \cellcolor{blue!25}99\% & \cellcolor{blue!25}99\% \\
          & Std  & 18\% & 13\% & 20\% & \cellcolor{orange!25}0.2\%& 1\%\\
\hline
60 / 1000 & Mean & 56\% & 85\% & 84\% & 97\%& \cellcolor{green!25}98\%\\
          & Max  & 92\% & 98\% & \cellcolor{blue!25}99\% & 98\%& \cellcolor{blue!25}99\%\\
          & Std  & 14\% & 17\% & 16\% & 1\%& \cellcolor{orange!25}0.5\%\\
\hline
60 / 1100 & Mean & 52\% & 59\% & 82\% & 60\%& \cellcolor{green!25}93\%\\
          & Max  & 79\% & 80\% & \cellcolor{blue!25}95\% & 94\%& \cellcolor{blue!25}95\%\\
          & Std  & 15\% & 17\% & 12\% & 18\%& \cellcolor{orange!25}1.5\%\\
\hline
70 / 900  & Mean & 59\% & 83\% & 81\% & 89\%& \cellcolor{green!25}96\%\\
          & Max  & 92\% & 98\%& 98\%& \cellcolor{blue!25}99\%& 98\%\\
          & Std  & 13\% & 14\% & 17\% & 13\%& \cellcolor{orange!25}2\%\\
\hline
70 / 1000 & Mean & 62\% & 76\% & 78\% & 61\%& \cellcolor{green!25}97\%\\
          & Max  & 84\% & 93\% & \cellcolor{blue!25}99\% & 70\%& \cellcolor{blue!25}99\%\\
          & Std  & 12\% & 18\% & 18\% & 4\%& \cellcolor{orange!25}1.5\%\\
\hline
70 / 1100 & Mean & 59\% & 65\% & 74\% & 61\%& \cellcolor{green!25}89\%\\
          & Max  & 84\% & 84\% & \cellcolor{blue!25}90\%& 70\%& \cellcolor{blue!25}90\%\\
          & Std  & 14\% & 14\% & 17\% & 5\%& \cellcolor{orange!25}1.5\%\\
\hline
\end{tabular}

\caption{Performance comparison of all methods. The best mean (green), best maximum (blue), and lowest standard deviation (orange) are highlighted. Among all methods, NOTES achieves the best performance, consistently producing high-performance, fully binarized designs while exhibiting the lowest variance, indicating reliable convergence to high-quality optima.}
\label{table:results}
\end{table*}

\section{Discussion}

Our framework imposes inductive biases in the neural reparameterization through the use of physics-informed designs, the latent-space representation, and the neural operator. This reparameterization scheme enables optimization to take place in a fully unconstrained space while satisfying binarization constraints. Currently, NOTES implements the binarization constraint through the neural operator. For more complex devices, rigorous methods have been developed to enforce fabrication constraints~\cite{chen2024validation, terekhov2024enhancing, hiesener2025seeded}. Most designs generated by the neural operator are already fully binary when coupled with CMA-ES. However, when coupled with gradient-based methods, the DeepONet often struggles to produce binarized designs. Although the exact cause requires further investigation, this difficulty may be dependent on the optimization formulation~\cite{bertsekas1997nonlinear, nocedal2006numerical}. The latent space induced by the neural operator does not inherently eliminate poor local optima. Consequently, gradient-based methods may be trapped in poor local minima, whereas CMA-ES is more robust and consistent in a low-dimensional space.

Our physics-informed ablation study suggests that dimensionality reduction inherently restricts the feasible design manifold explored during optimization. Consequently, incorporating a physics-informed prior is critically important for guiding the search toward high-performance regions of the design space. The additional experiment with PCA decoder shows that the neural operator learns a better non-linear mapping to a more complex design manifold in high-dimensional space than a linear decoder. NOTES outperforms its counterparts in ablation studies for the 2D structural optimization problem, whereas the performance gap is smaller in the simpler 1D case, suggesting that the benefits from NOTES increase with the problem's complexity. Although NOTES transfers effectively across inverse design conditions and parameters within the same PDE family, the limits of representation transferability across substantially different topology classes or physical regimes remain an open question.

\section{Conclusion}

Overall, NOTES establishes an effective physics-informed evolutionary strategy for generating high-performance and physically realistic designs. By combining dimensionality reduction, neural operator representations, and CMA-ES, the framework achieves rapid and consistent convergence. Our results demonstrate that DeepONet is an effective nonlinear decoder that captures high-performance regions of topology manifolds. NOTES also demonstrates strong reusability across different resolutions, optimization constraints, and PDE parameters. Finally, we find that CMA-ES is more effective than L-BFGS at escaping poor local optima within the learned latent space.

In this work, NOTES is demonstrated using PCA as the dimensionality reduction method. In future work, other latent-space representations may be explored, including neural encoder-based approaches~\cite{bengio2013representation, lecun2015deep, marzban2025hilab}. Different latent-space representations may affect convergence stability and computational cost. Although gradient-based optimization coupled with neural representations (L-BFGS + DeepONet) achieves promising design performance and lower computational cost, the resulting designs are often not fully binary. Future work will explore subpixel-smoothed projection~\cite{hammond2025unifying} as a hard binarization constraint to improve the physical realism of gradient-based designs while mitigating the ill-conditioning associated with the hard thresholding of fabrication constraints.

\section{Data and Code Availability}

The results of this work used the \href{http://metanet.stanford.edu/search/dielectric-metagratings/}{Metanet}~\cite{jiang2020metanet} dataset.

The implementation of the MBB beam application follows the same framework as the prior work~\cite{hoyer2019neural}, but instead of TensorFlow, we implement the code in PyTorch.

The code implementation is available on \href{https://github.com/xmh3698/notes-pde-opt.git}{GitHub}.

\section{Acknowledgment}

X. Huang and R. Pestourie acknowledge support from the U.S. Department of Energy, Office of Science, Office of Biological and Environmental Research, U.S. National Science Foundation under Grants No.~IIS~2435905, and National Institute of Biomedical Imaging and Bioengineering of the National Institutes of Health under Award Number R21EB036343. G. Zhang acknowledges support from the U.S. Department of Energy, Office of Science, Office of Advanced Scientific Computing Research, Applied Mathematics program, under the contracts ERKJ388 and ERKJ456 at Oak Ridge National Laboratory. L.Lu was supported by the U.S. DOE ASCR under Grants No.~DE-SC0025593 and No.~DE-SC0025592, and the U.S. National Science Foundation under Grants No.~DMS-2347833 and No.~DMS-2527294.

\bibliographystyle{unsrt}
\bibliography{reference}

\end{document}


\title{Supplementary Information}
\date{}
\maketitle


In the following sections, we provide additional figures, ablation studies, and reproducible experimental details for NOTES applied to both the nanophotonic beam deflector problem and the MBB beam structural optimization problem. The nanophotonic application section (Section~\ref{si-sec:nano}) includes convergence-rate comparisons with standard CMA-ES, DeepONet training curves, reconstruction quality, latent-space gradient analyses, additional optimization experiments, and ablation studies comparing DeepONet and PCA-decoder as well as physics-informed and randomly sampled training data. The MBB beam application section (Section~\ref{si-sec:2D}) presents reproducible details for data generation and processing, PCA dimensionality reduction, DeepONet architecture and training, optimization settings, transfer-learning experiments across resolutions and volume fractions, and additional ablation studies comparing NOTES against PCA-decoder, randomly sampled training data, and L-BFGS + DeepONet baseline. These supplementary results further support the robustness and reproducibility of the proposed NOTES framework.

\section{Additional Details for Nanophotonic application}
\label{si-sec:nano}

\subsection{Convergence rate comparison}
\label{si-subsec:converge}

\begin{figure}[H]
    \centering

    \subfloat[Example 1]{%
        \includegraphics[width=0.48\linewidth]{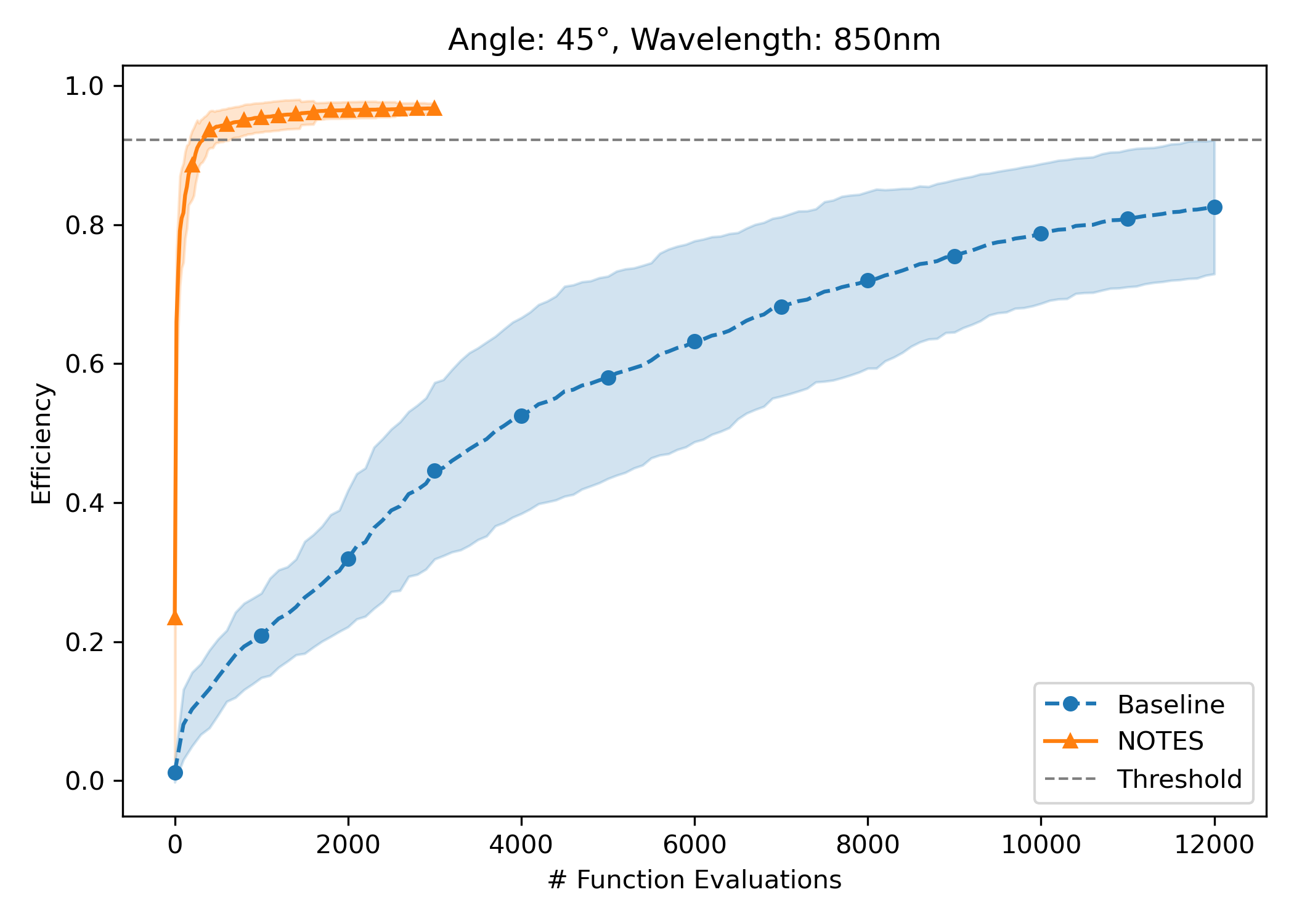}
    }
    \hfill
    \subfloat[Example 2]{%
        \includegraphics[width=0.48\linewidth]{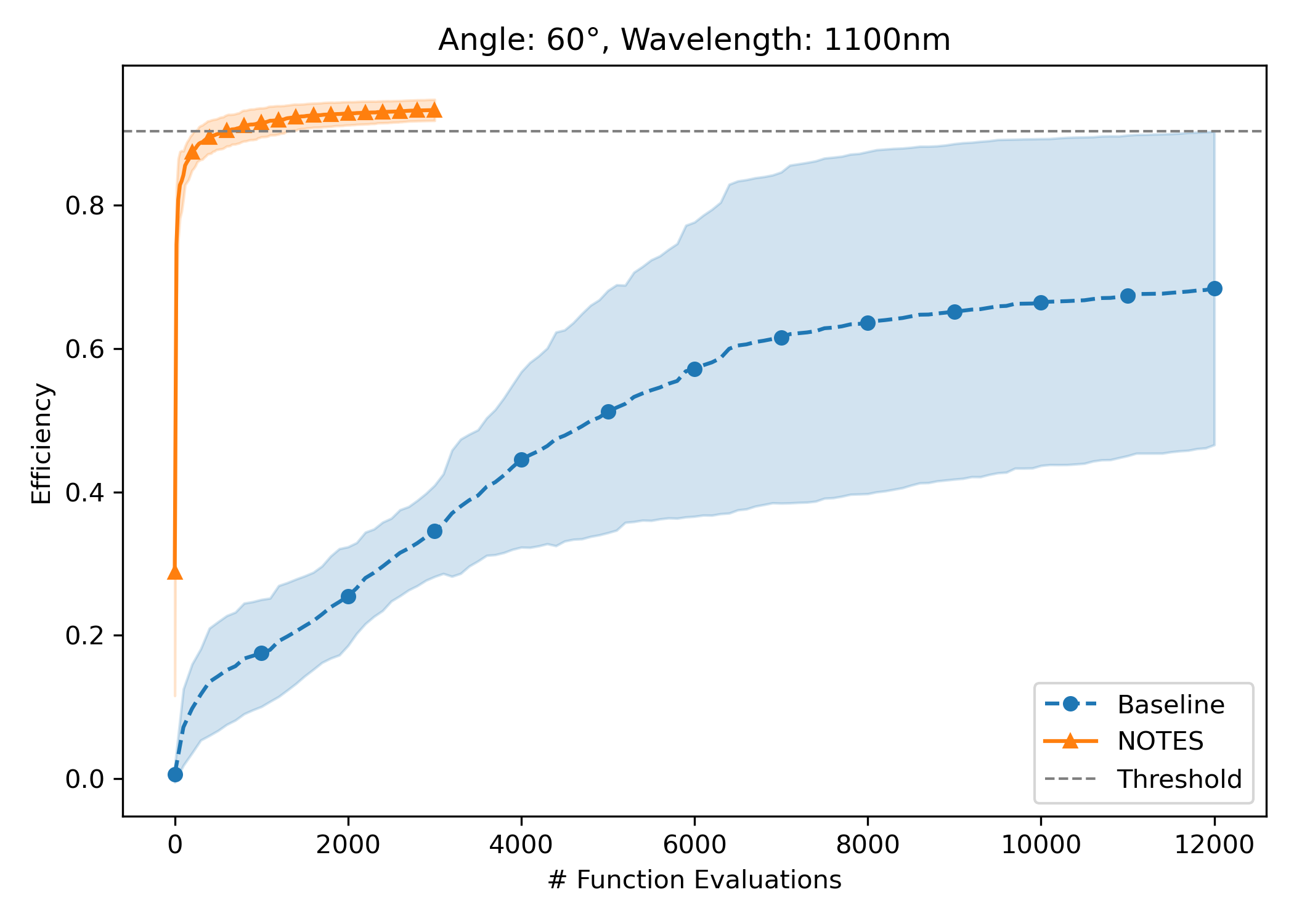}
    }

    \caption{Comparison of convergence histories of NOTES (orange) and standard CMA-ES (blue). The in-figure legend entry “Baseline” refers to standard CMA-ES. The threshold, shown as a dashed gray horizontal line, is defined as one standard deviation above the mean value of standard CMA-ES at the last epoch. The curves represent mean value at each epoch. The upper and lower bounds indicate one standard deviation above and below the mean, respectively. Each dot on the curves corresponds to an interval of 10 epochs.}
    \label{fig:compare_notes_cma}
\end{figure}

Figure~\ref{fig:compare_notes_cma} shows the comparison between NOTES and standard CMA-ES. NOTES optimizes over a 25-dimensional space, while standard CMA-ES optimizes over a 256-dimensional space. Each method applies 30 runs, and the best-so-far objective values are plotted against the number of function evaluations. We observe that NOTES achieves a faster convergence rate than CMA-ES due to the reduced degrees of freedom.

\subsection{Training curve}
\label{si-subsec:1d-train}

\begin{figure}[H]
    \centering
    \includegraphics[width=1\linewidth]{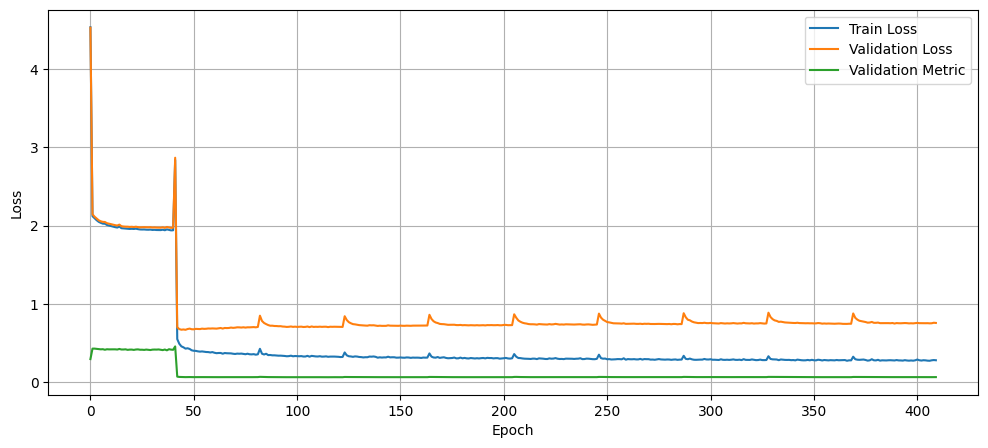}
    \caption{Training loss is shown by the blue curve, while validation loss is shown by the orange curve. An additional validation metric, mean squared error (MSE), is shown by the green curve.}
    \label{fig:deeponet-1d-loss}
\end{figure}

Figure~\ref{fig:deeponet-1d-loss} shows the loss curve of DeepONet trained for 400,000 epochs. Every 40,000 epochs, the scale factor of the sigmoid function in the DeepONet is doubled. Although the gap between the training loss and validation loss increases over the epochs, the validation metric, mean squared error (MSE), remains approximately unchanged at around 0.06.

\subsection{Reconstruction Quality}
\label{si-subsec:1d-reconstruct}

\begin{figure}[H]
    \centering

    \subfloat[Sample 1\label{fig:reconstruction-ex1}]{
        \includegraphics[width=0.48\linewidth]{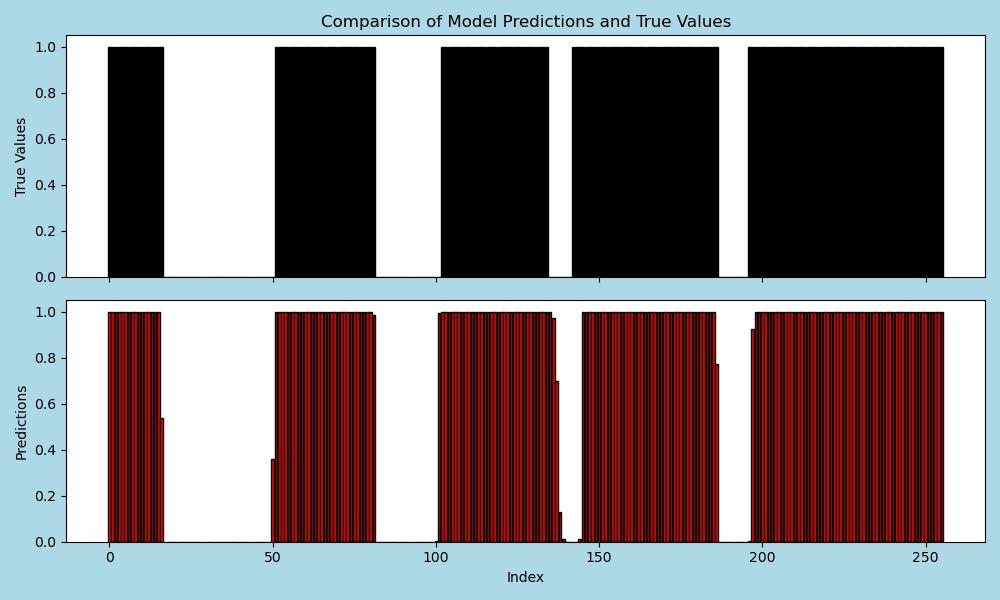}
    }
    \hfill
    \subfloat[Sample 2\label{fig:reconstruction-ex2}]{
        \includegraphics[width=0.48\linewidth]{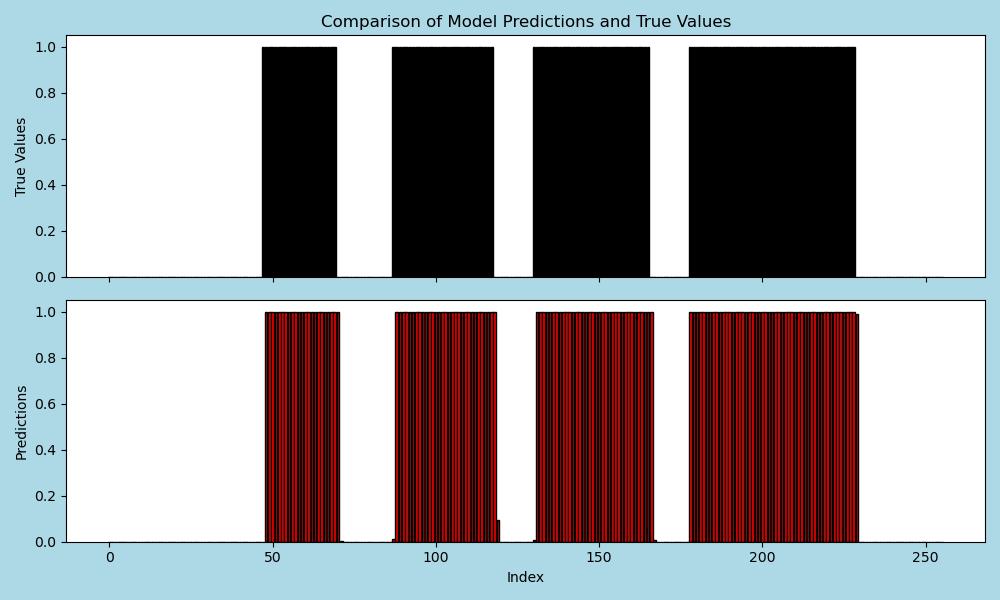}
    }

    \vspace{0.5cm}

    \subfloat[Sample 3\label{fig:reconstruction-ex3}]{
        \includegraphics[width=0.48\linewidth]{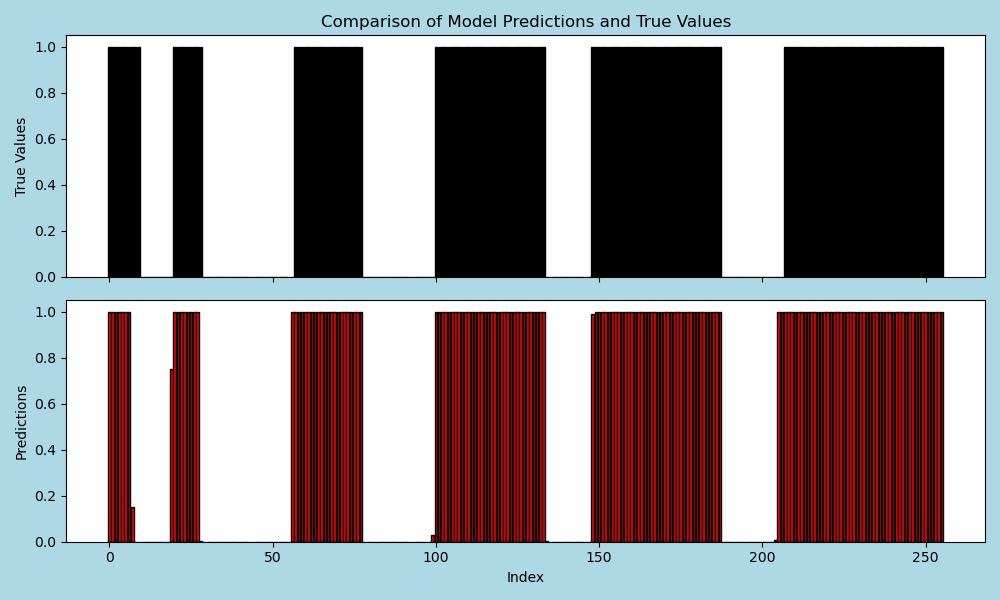}
    }
    \hfill
    \subfloat[Sample 4\label{fig:reconstruction-ex4}]{
        \includegraphics[width=0.48\linewidth]{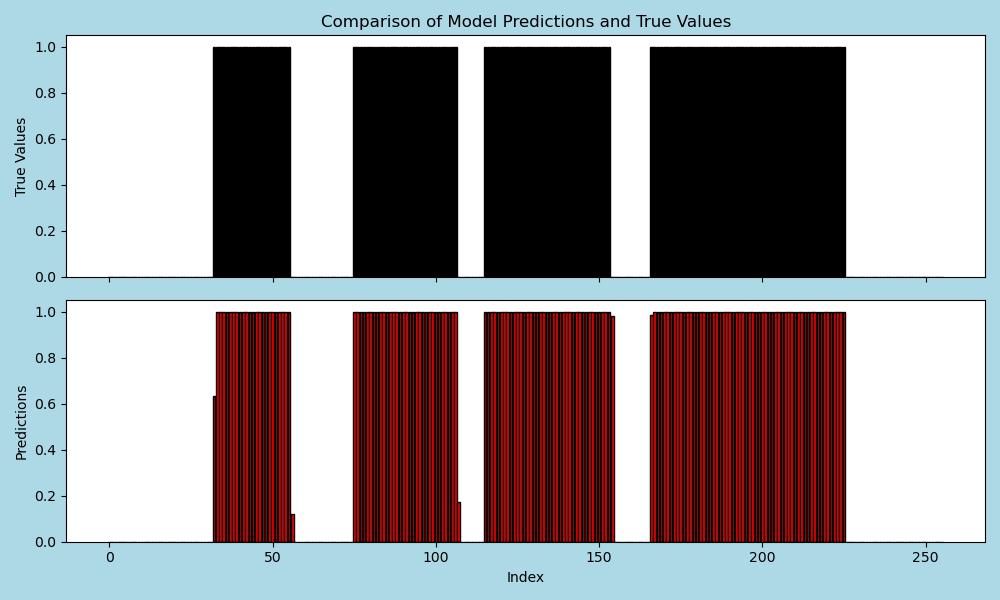}
    }

    \caption{Visualization of the reconstruction quality of DeepONet for the nanophotonic application. The upper panel shows the ground-truth design (black), while the lower panel shows the corresponding design reconstructed by DeepONet (red).}
    \label{fig:reconstruction-examples}
\end{figure}

Figure~\ref{fig:reconstruction-examples} shows the reconstruction quality of DeepONet. Overall, DeepONet predicts the locations and widths of the ridges in each design quite well, but some ridge edges are not sharp.

\subsection{Additional Gradient Descent}
\label{si-subsec:1d-gd}


To verify whether the solutions produced by L-BFGS + DeepONet are trapped near local optima, we performed an additional gradient descent refinement experiment. Specifically, we apply 2,000 iterations of gradient descent with a learning rate of $10^{-3}$ to the latent variables initialized from the L-BFGS + DeepONet solutions. The learning rate was selected based on preliminary gradient-norm-versus-iteration tests, in which larger learning rates resulted in unstable optimization trajectories. If the L-BFGS + DeepONet solutions are not near stationary points, this additional refinement should substantially reduce both the latent gradient norms and the corresponding objective values. In this experiment, we considered 100 optimized designs at a deflection angle of $40^\circ$ and a wavelength of $1000$~nm, and compared the latent-space gradient norms before and after refinement.

\begin{figure}[H]
    \centering

    \subfloat[CDF of Gradient Norm\label{fig:CDF-grad-norm}]{
        \includegraphics[width=0.48\linewidth]{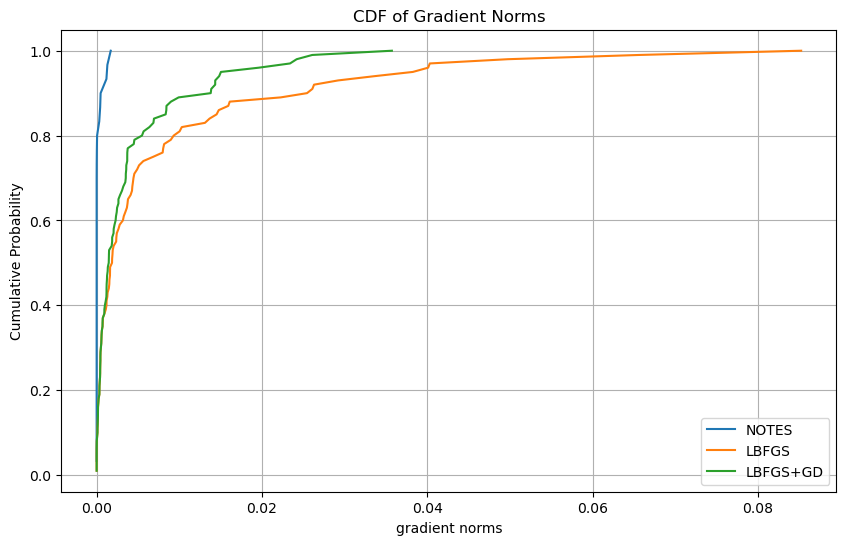}
    }
    \hfill
    \subfloat[Bar plot of Gradient Norm\label{fig:bar-grad-norm}]{
        \includegraphics[width=0.48\linewidth]{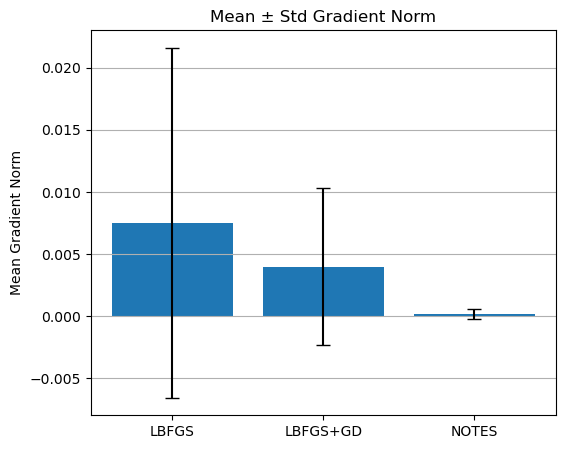}
    }

    \caption{Comparison of latent-space gradient norms for NOTES, L-BFGS + DeepONet, and L-BFGS + DeepONet followed by gradient descent.}
    \label{fig:grad-norm-comparison}
\end{figure}

The comparison of latent-space gradient norms (Figure~\ref{fig:grad-norm-comparison}) shows that additional gradient descent consistently reduces the gradient magnitude of the L-BFGS + DeepONet solutions. The average gradient norm decreases from $7.5 \times 10^{-3}$ before gradient descent to $4.0 \times 10^{-3}$ after gradient descent, while the standard deviation decreases from $1.41 \times 10^{-2}$ to $6.3 \times 10^{-3}$.

\begin{figure}[H]
    \centering

    \subfloat[CDF of Efficiency\label{fig:cdf-efficiency}]{
        \includegraphics[width=0.48\linewidth]{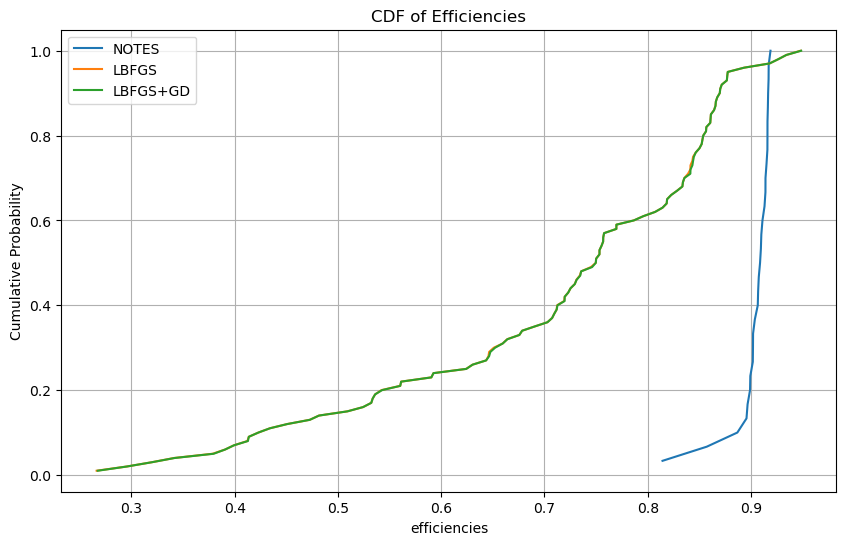}
    }
    \hfill
    \subfloat[Bar plot of Efficiency\label{fig:bar-efficiency}]{
        \includegraphics[width=0.48\linewidth]{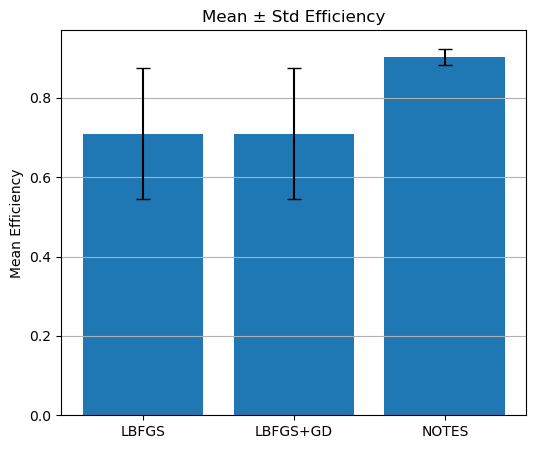}
    }

    \caption{Comparison of the final objective values for designs generated by NOTES, L-BFGS + DeepONet, and L-BFGS + DeepONet followed by gradient descent. The curves for L-BFGS + DeepONet and L-BFGS + DeepONet followed by gradient descent overlap almost completely.}
    \label{fig:efficiency-comparison}
\end{figure}

In contrast, the comparison of deflection efficiency (Figure~\ref{fig:efficiency-comparison}) shows that the efficiency distribution remains nearly unchanged before and after the additional gradient descent refinement. Although the latent-space gradient norms are reduced, the objective values exhibit no meaningful improvement. This indicates that the additional gradient descent steps only move the solutions within the same attraction basin without escaping to significantly better designs. Therefore, the solutions produced by L-BFGS + DeepONet are already trapped near poor local optima in the latent landscape.

\subsection{DeepONet vs PCA-decoder}
\label{si-subsec:1d-pca}

For this ablation study, we replace the DeepONet decoder in NOTES with a linear PCA decoder for the 1D nanophotonic application. We use the same dataset employed to train the DeepONet, consisting of all designs with precomputed deflection efficiencies greater than 0.9, to derive the PCA representation. Specifically, PCA decomposition is applied to obtain the first 25 principal components together with the mean vector. The PCA decoder reconstructs the designs by reversing the PCA projection using these retained components.

\begin{align}
    \sigma(x;\beta,\eta) &= \frac{1}{1+\exp(-\beta x-\eta)} \label{custom_sig}
\end{align}

\begin{figure}[H]
    \centering
    \includegraphics[width=0.5\linewidth]{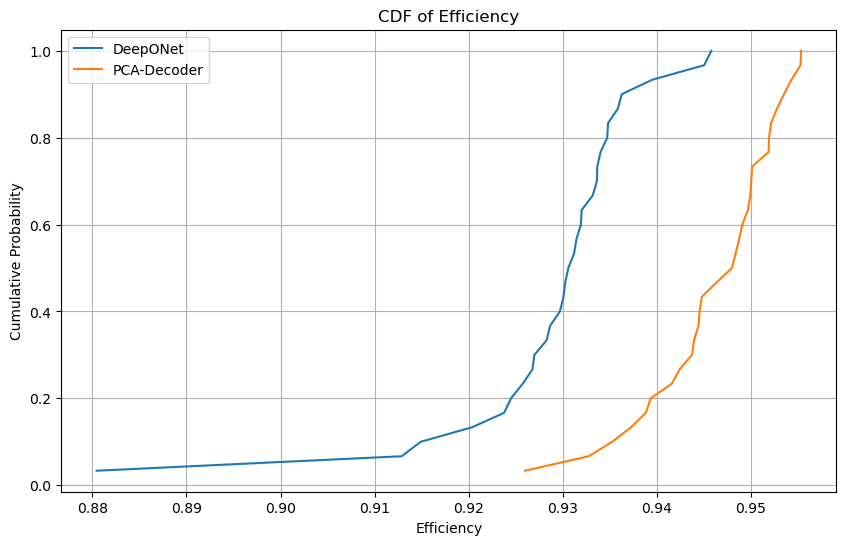}
    \caption{Comparison of the final efficiency for the PCA decoder and DeepONet, both combined with CMA-ES, at a wavelength of 1100 and an angle of 60.}
    \label{fig:1d-pca-decoder}
\end{figure}

To enforce optimization constraints, we apply the same custom sigmoid function~\ref{custom_sig} used in NOTES, with a scale factor of 10 and a threshold of 0. The optimization framework is otherwise identical to NOTES, except that the DeepONet decoder is replaced by the PCA decoder. CMA-ES is initialized from a unit Gaussian distribution in the latent space with an initial step size of $\sigma_0 = 2.0$. As shown in Figure \ref{fig:1d-pca-decoder}, CMA-ES combined with the PCA decoder achieves slightly better performance than NOTES. Both DeepONet and PCA-decoder are consistent at generating high-performance designs.

\subsection{Randomly-Sampled vs Physics-Informed data}
\label{si-subsec:1d-ab-pi}

To further validate that the performance of NOTES arises from its physics-informed training data rather than from the DeepONet architecture alone, we conduct an ablation study on the nanophotonic application using randomly sampled designs. Specifically, we generate a dataset of the same size as the physics-informed training dataset by independently sampling each pixel from a Bernoulli distribution. A new DeepONet is trained on this randomly sampled dataset using exactly the same training procedure and architecture as the physics-informed counterpart.

The results indicate that NOTES trained on randomly generated designs performs worse than NOTES trained on physics-informed designs. Although both methods achieve similar maximum efficiencies (approximately 94–95\%), the randomly initialized approach exhibits significantly larger variance in performance. The variance in efficiency is approximately one order of magnitude larger than that of the physics-informed counterpart. In addition, more than 75\% of the designs generated from random initialization have aspect ratios greater than 33, whereas fewer than 5\% of the physics-informed designs exceed this threshold. These results indicate that the physics-informed prior substantially improves optimization consistency. Consequently, obtaining high-performance designs using randomly sampled data would require significantly higher computational cost.

\begin{figure}[H]
    \centering

    \subfloat[Distribution of final efficiency for physics-informed data\label{fig:hist-eff-pi}]{
        \includegraphics[width=0.48\linewidth]{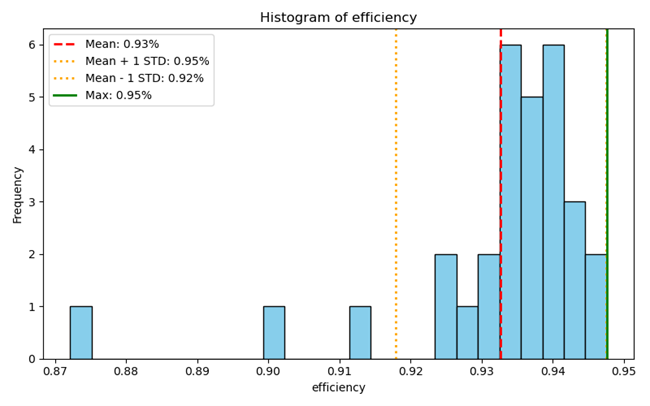}
    }
    \hfill
    \subfloat[CDF of aspect ratios for physics-informed data\label{fig:aspect-ratio-pi}]{
        \includegraphics[width=0.48\linewidth]{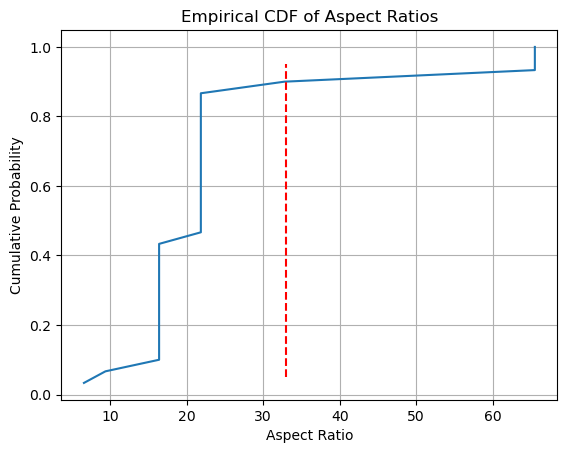}
    }

    \vspace{0.5cm}

    \subfloat[Distribution of final efficiency for randomly-sampled data\label{fig:hist-eff-rand}]{
        \includegraphics[width=0.48\linewidth]{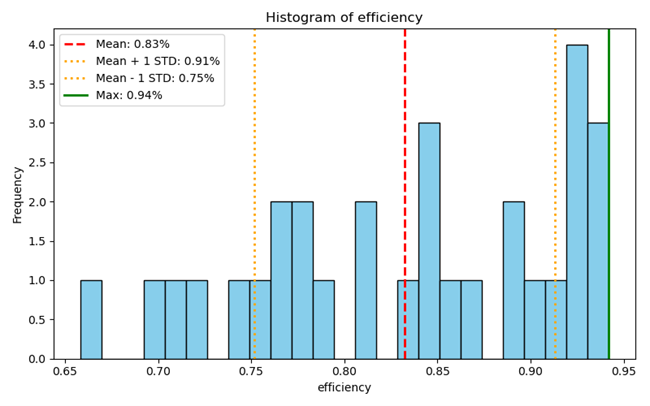}
    }
    \hfill
    \subfloat[CDF of aspect ratios for randomly-sampled data\label{fig:aspect-ratio-rand}]{
        \includegraphics[width=0.48\linewidth]{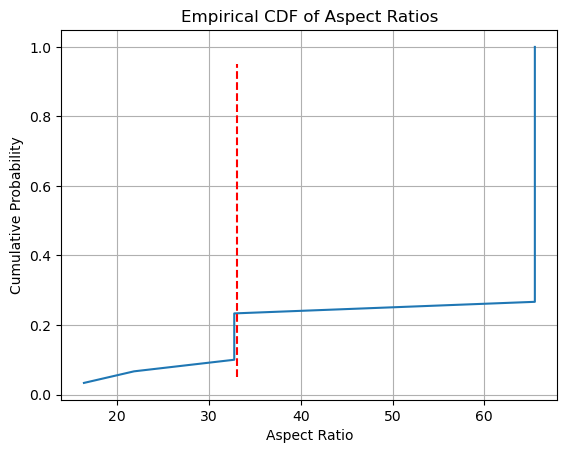}
    }

    \caption{Comparison between physics-informed and randomly sampled training data. The red dashed vertical line in the CDF plots highlights the threshold used to distinguish designs with low and high aspect ratios.}
    \label{fig:physics-informed-vs-random}
\end{figure}

\section{Experiment details for Structural Optimization}
\label{si-sec:2D}

The second application is a structural optimization problem. We apply NOTES to Messerschmitt-Bölkow-Blohm (MBB) beams at a resolution of $192 \times 64$, which corresponds to 12,288 pixels. Using PCA, the latent space is reduced to 60 dimensions. By training DeepONet with augmented data, NOTES can discover high-performance designs. Moreover, when the trained DeepONet is transferred to optimize MBB beams with different resolutions and volume constraints, the method still produces competitive results.

\subsection{Data Generation}
\label{si-subsec:generation}




The training data is generated using topology optimization with L-BFGS optimizer. The volume constraint is enforced using a custom sigmoid function with a threshold derived via binary search. The initial design is set such that all entries are equal to the volume constraint, a predefined value between 0 and 1. After optimization is completed, we randomly erode the optimized design for 1 to 8 iterations. We repeat this process to generate 1,000 designs.

\subsection{Data Processing}
\label{si-subsec:processing}

\begin{figure}[H]
    \centering
    \includegraphics[width=0.5\linewidth]{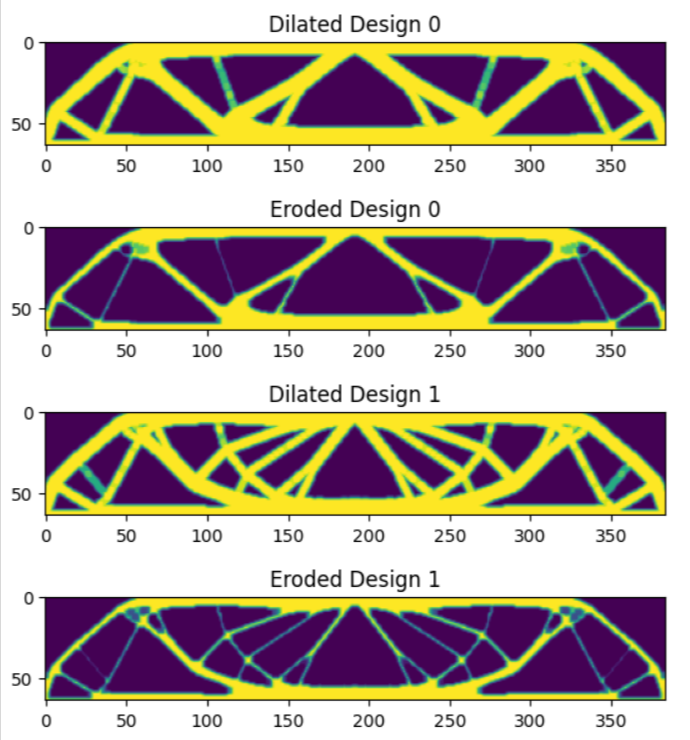}
    \caption{Examples of eroded and dilated designs.}
    \label{fig:augmentation}
\end{figure}

\begin{figure}[H]
    \centering

    \subfloat[PCA explained variance\label{fig:pca-explained-variance}]{
        \includegraphics[width=0.48\linewidth]{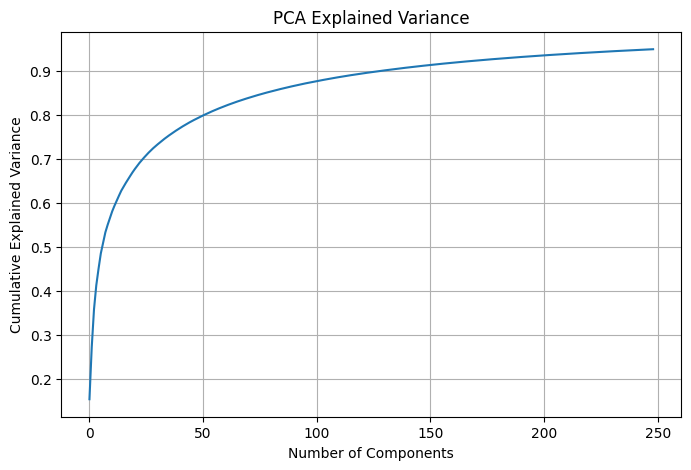}
    }
    \hfill
    \subfloat[Visualization of designs in the first two principal components\label{fig:pca-projection}]{
        \includegraphics[width=0.48\linewidth]{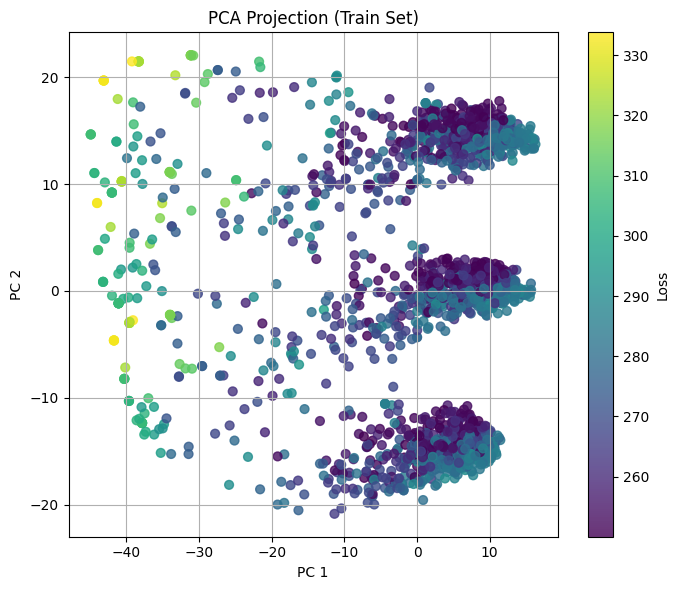}
    }

    \caption{PCA analysis of the eroded, dilated, and original designs.}
    \label{fig:pca-analysis}
\end{figure}

The median compliance of the designs is about 270. We generate designs with compliance greater than 250 as training data. There are a total of 857 such designs. For each design, we apply dilation and erosion to obtain two augmented designs. In total, there are 2056 designs for training and 515 designs for validation. Figure~\ref{fig:augmentation} shows designs after dilation and erosion. We apply PCA and keep the first 60 principal components to form the latent space. According to Figure~\ref{fig:pca-explained-variance}, these components explain about 83\% of the variance in the designs. Qualitatively, the dataset projected onto the first and second principal components shows clear clustering (Figure~\ref{fig:pca-projection}), indicating that the dataset may have exploitable low dimensionality. Before the training of DeepONet, the PCA coordinates are normalized to have zero-mean and unit standard deviation along each dimension. This additional step simplifies the selection of the sampling radius for the downstream CMA-ES optimization.

\subsection{Model Training}
\label{si-subsec:2d-train}

\begin{figure}[H]
    \centering
    \includegraphics[width=0.5\linewidth]{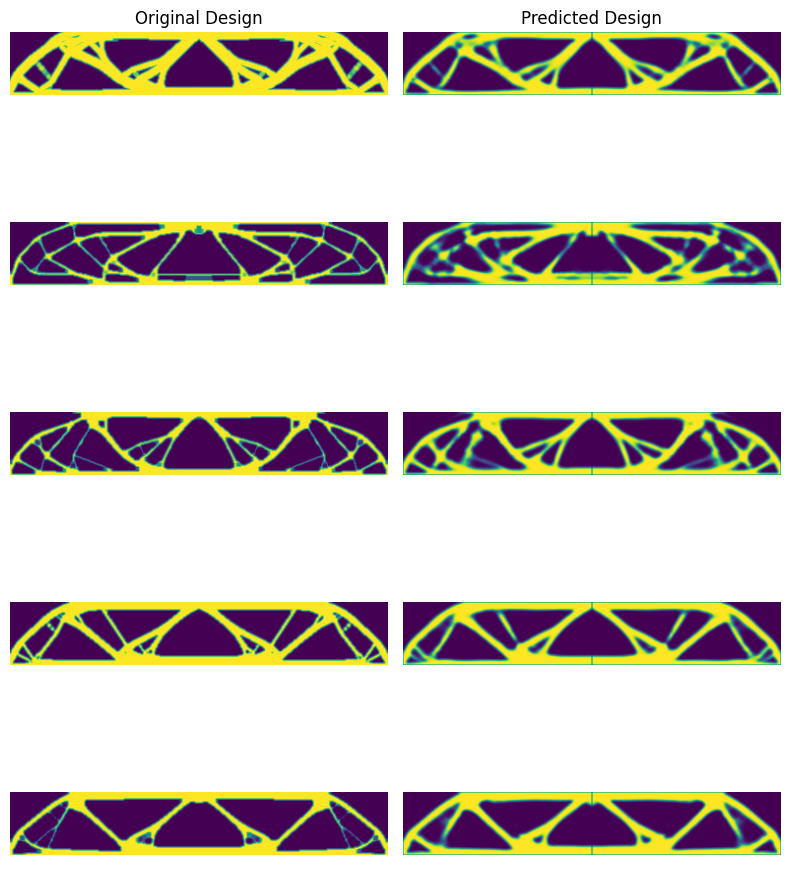}
    \caption{Reconstruction quality of DeepONet for the 2D structural optimization application. The left column shows the eroded, dilated, and original designs, while the right column shows the corresponding designs reconstructed by DeepONet.}
    \label{fig:2d-reconstruction}
\end{figure}

A DeepONet with 1.6M parameters is trained using this training data. Both the trunk and branch networks consist of four fully connected hidden layers with 512 neurons per layer and ReLU activation functions. The network weights are initialized using Glorot normal initialization. We apply a custom sigmoid function after the last layer of DeepONet to ensure that the outputs are between 0 and 1. The loss function is binary cross-entropy loss, with an additional binary soft-constraint term added to the objective. DeepONet is trained for 30K epochs. Figure~\ref{fig:2d-reconstruction} shows several comparisons between the true designs and the predicted designs. The reconstruction quality demonstrates that DeepONet captures large geometric features of the designs quite well.

\subsection{Optimization}
\label{si-subsec:2d-opt}

During optimization, we replace the custom sigmoid with the hard constraint transformation mentioned in main text (Section V-C). We apply NOTES with a computational budget capped at 2.4k objective function evaluations. The population size is set to 30, and the initial sampling radius is set to 2.0. The best design has a compliance of 246. It is worth noting that the training data consisted of segmented designs with compliance greater than 250. Yet, DeepONet learns geometric features that achieve a design that is 1.6\% better than the best design in the training data.

\begin{figure}[H]
    \centering
    \includegraphics[width=0.5\linewidth]{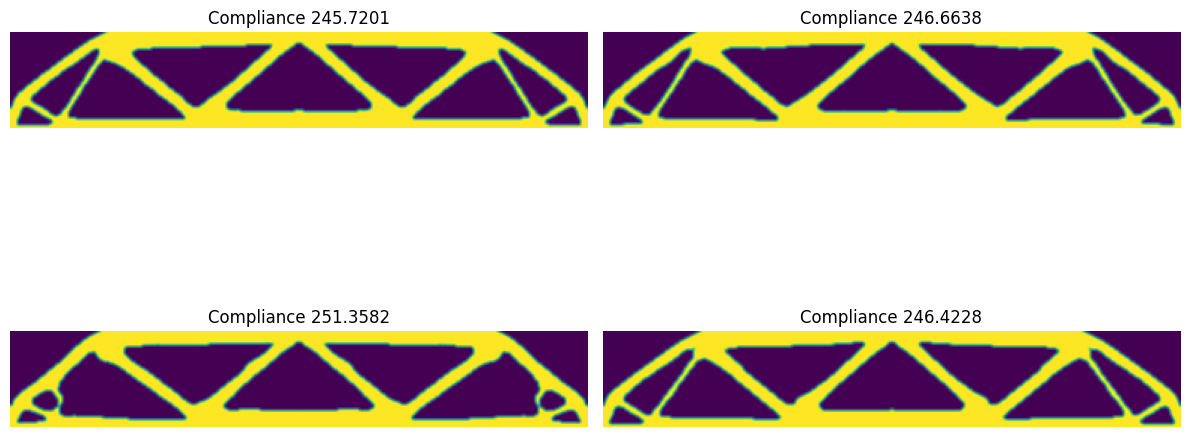}
    \caption{Optimization results generated by NOTES.}
    \label{fig:2d-opt}
\end{figure}

As Figure \ref{fig:2d-opt} shows, we observe mode collapse as most optimized designs look very similar. It may be due to an imbalance of geometric features of the training data. Many designs in the training data share the same shape at the center and have an L-shape support on the wing.

\begin{figure}[H]
    \centering
    \includegraphics[width=0.5\linewidth]{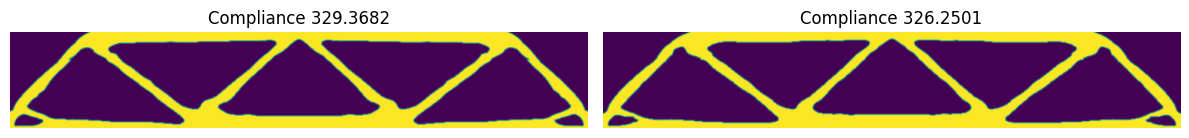}
    \caption{Transfer learning for higher resolution, from 192 $\times$ 64 to 384 $\times$ 128.}
    \label{fig:high-res}
\end{figure}

\begin{figure}[H]
    \centering
    \includegraphics[width=0.5\linewidth]{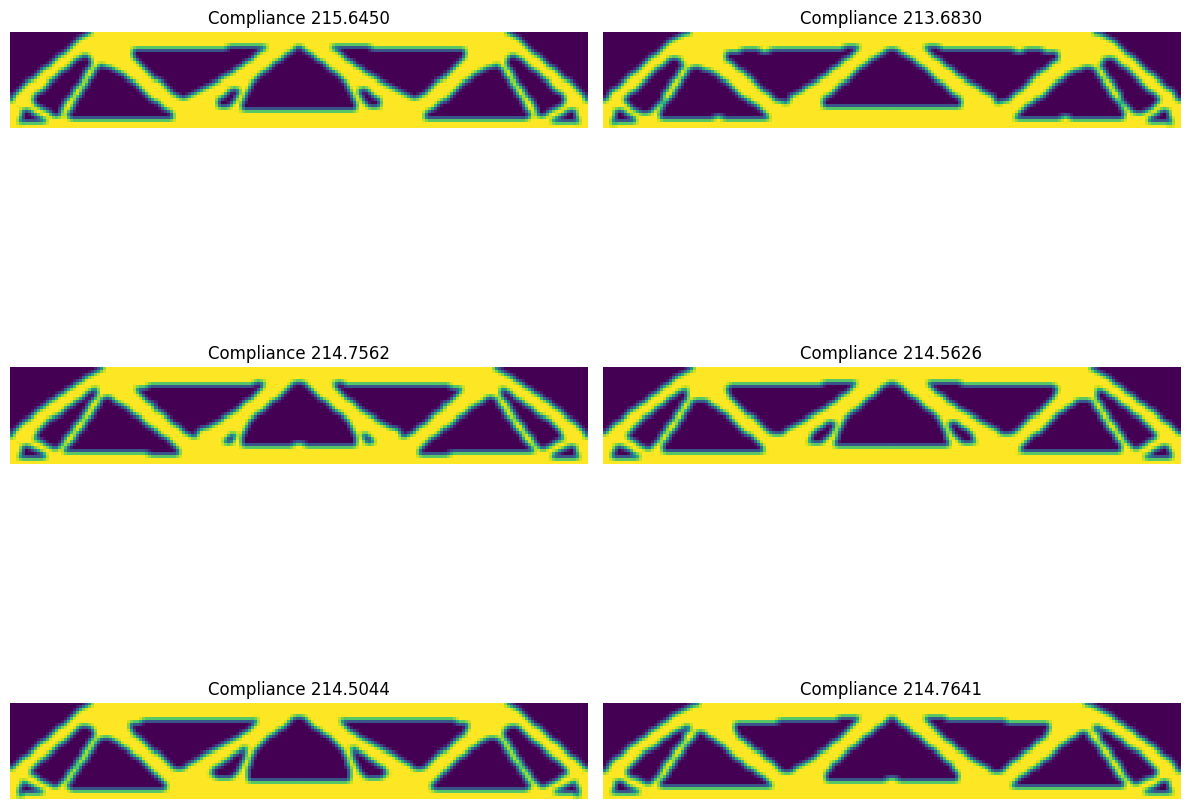}
    \caption{Transfer learning for lower resolution, from 192 $\times$ 64 to 96 $\times$ 32.}
    \label{fig:low-res}
\end{figure}

We reuse the pretrained DeepONet to optimize designs at a coarser resolution of $96\times32$ with a volume constraint of 0.5, or at a finer resolution of $384\times128$ with a volume constraint of 0.3. As shown in Figure \ref{fig:low-res} and Figure \ref{fig:high-res}, we observe competitive performance without retraining. Table \ref{tab:opt-result} summarizes the optimization results. It indicates that the one-shot multifidelity approach is particularly beneficial for the finer resolution, where creating a dataset would have been one or two orders of magnitude more computationally expensive, while having slightly worse performance than results of pixel L-BFGS baseline from prior work\cite{hoyer2019neural}.

\begin{table}[H]
    \centering
    \begin{tabular}{c|c|c|c}
    \hline
         & 96 $\times$ 32, 0.5 &192 $\times$ 64, 0.4 &384 $\times$ 128, 0.3\\
     \hline
    Best design & 214 & 246 & 326\\
    \hline
    Improvement against & -2.4\% & 1.6\% & -1.8\%\\
    pixel L-BFGS & & & \\
    \hline
    \end{tabular}
    \caption{Optimization Results}
    \label{tab:opt-result}
\end{table}

\subsection{DeepONet vs PCA-decoder for 2D}
\label{si-subsec:2d-pca}

For the PCA-decoder ablation study on the MBB beam application, we use the same training dataset for DeepONet. The PCA projection matrix and mean vector are obtained through flattened designs. The PCA decoder reconstructs the designs by reversing the PCA projection, adding the mean vector, and reshaping back to original shape. It replaces the DeepONet decoder within the NOTES optimization framework. CMA-ES is initialized from a unit Gaussian distribution in the latent space with an initial step size $\sigma_0 = 2.0$.

\begin{figure}[H]
    \centering

    \subfloat[Design produced by CMA-ES + PCA with the same computational cost\label{fig:pca-same-comp}]{
        \includegraphics[width=0.48\linewidth]{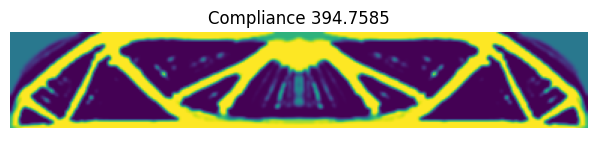}
    }
    \hfill
    \subfloat[Design produced by CMA-ES + PCA with doubled computational cost\label{fig:pca-double-comp}]{
        \includegraphics[width=0.48\linewidth]{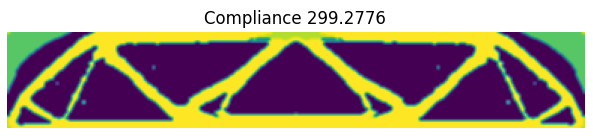}
    }

    \caption{Comparison of designs generated by CMA-ES + PCA under different computational budgets.}
    \label{fig:pca-compute-budget}
\end{figure}

For the geometrically more complex 2D structural optimization problem, the PCA decoder substantially underperforms the proposed NOTES framework, as shown in Figure \ref{fig:pca-compute-budget}. Under the same computational budget, NOTES identifies designs with compliance as low as 246 for the $192 \times 64$ MBB beam case, whereas CMA-ES coupled with the PCA decoder only achieves a best compliance of approximately 394. Even when the PCA-decoder baseline is allocated roughly twice the computational budget of NOTES, the best compliance only improves to around 300, after which no meaningful further improvement is observed. These results suggest that the nonlinear design manifold learned by DeepONet is significantly more expressive than a linear PCA representation for high-dimensional PDE-constrained structural optimization.

\subsection{Randomly-sampled vs Physics-informed data in 2D}
\label{si-subsec:2d-ab-pi}

We further investigate the importance of the physics-informed component in NOTES through an ablation study on the MBB beam structural optimization problem.

We generate random design logits with the same spatial resolution and sample size as the L-BFGS-generated dataset. Each sample is then processed using the volume constraint function, followed by cone filtering to enforce the prescribed volume fraction and to eliminate isolated small-scale features. This yields a total of 2,571 training samples for the ablation study.

\begin{figure}[H]
    \centering
    \includegraphics[width=0.5\linewidth]{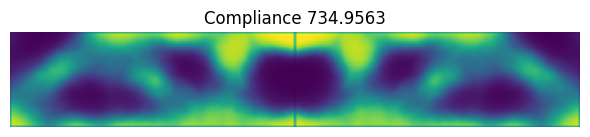}
    \caption{Example design when NOTES is not physics-informed}
    \label{fig:ab-no-pi}
\end{figure}

A DeepONet with the identical architecture used in NOTES is then trained on this synthetic dataset. At convergence, the binary cross-entropy loss remains at 0.675. For comparison, DeepONet trained with physics-informed designs has a binary cross-entropy loss around 0.1. We then apply this trained DeepONet with CMA-ES to optimize with the same procedure as NOTES. In the best run, the initial compliance is 2466. After 80 iterations and 2400 function evaluations, the resulting compliance is 735, as shown in Figure \ref{fig:ab-no-pi}.

\subsection{L-BFGS + DeepONet vs NOTES in 2D}
\label{si-subsec:2d-lbfgs}

We apply L-BFGS + DeepONet to the 2D MBB beam structural optimization problem. Unlike the nanophotonic application, the volume constraint is explicitly enforced within the optimization pipeline. Consequently, all generated designs with L-BFGS optimizer satisfy the prescribed volume constraint. The same DeepONet used in NOTES is coupled with L-BFGS optimizer in this experiment. For each run, the initial starting point in the latent space is uniformly sampled along each dimension between -1 and 1. The initial learning rate is set to 0.5, which was selected after testing learning rates ranging from 0.001 to 1.

\begin{figure}[H]
    \centering

    \subfloat[Example 1\label{fig:lbfgs-deeponet-ex1}]{
        \includegraphics[width=0.48\linewidth]{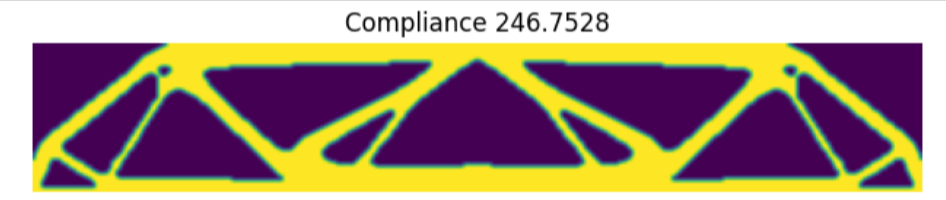}
    }
    \hfill
    \subfloat[Example 2\label{fig:lbfgs-deeponet-ex2}]{
        \includegraphics[width=0.48\linewidth]{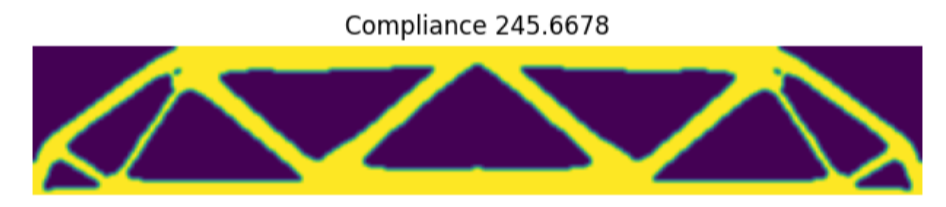}
    }

    \caption{Optimization results generated by L-BFGS + DeepONet.}
    \label{fig:lbfgs-deeponet-designs}
\end{figure}
 
As shown in Figure \ref{fig:lbfgs-deeponet-designs}, NOTES and L-BFGS + DeepONet achieve comparable performance. Both methods converge to designs with similar structures and identical compliance values. However, L-BFGS + DeepONet requires approximately 10 times fewer function evaluations, making it a highly competitive alternative for this optimization problem.

\bibliographystyle{unsrt}
\bibliography{reference}